%% file: iclr2026_conference.tex
\documentclass{article} 
\usepackage{iclr2026_conference,times}

\input{math_commands.tex}

\usepackage{hyperref}
\usepackage{url}
\usepackage{booktabs}
\usepackage{subcaption}
\usepackage{graphicx}
\usepackage{multirow}
\usepackage{array}
\usepackage{geometry}
\usepackage{float}
\usepackage{cleveref}
\usepackage{authblk}

\title{Data Scaling Laws for Radiology Foundation Models}

\author[1]{Maximilian Ilse*}
\author[1]{Harshita Sharma}
\author[1]{Anton Schwaighofer}
\author[1]{Sam Bond-Taylor}
\author[1]{Fernando P\'erez-Garc\'ia}
\author[2]{Olesya Melnichenko}
\author[3]{Anne-Marie G. Sykes, M.D.†}
\author[6]{Kelly K. Horst, M.D.†}
\author[3]{Ashish Khandelwal, M.D.†}
\author[3]{Maxwell Reynolds}
\author[1, 5]{Maria T. Wetscherek, M.D.}
\author[4]{Noel C. F. Codella}
\author[1]{Javier Alvarez-Valle}
\author[3]{Korfiatis Panagiotis*‡}
\author[1]{Valentina Salvatelli‡}

\affil[1]{Microsoft Health Futures UK}
\affil[2]{Microsoft Health Futures US}
\affil[3]{Mayo Clinic}
\affil[4]{Microsoft Health \& Life Sciences}
\affil[5]{Cambridge University Hospitals}
\affil[6]{Radiology AI Lab, Mayo Clinic}

\iclrfinalcopy
\begin{document}

\maketitle

\let\thefootnote\relax\footnotetext{* Corresponding authors \{ilse.maximilian, korfiatis.panagiotis\}@mayo.edu}
\footnotetext{† Equal clinical supervision}
\footnotetext{‡ Shared senior author}

\begin{abstract}
Foundation vision encoders such as CLIP and DINOv2, trained on web-scale data, exhibit strong transfer performance across tasks and datasets. However, medical imaging foundation models remain constrained by smaller datasets, limiting our understanding of how data scale and pretraining paradigms affect performance in this setting. In this work, we systematically study continual pretraining of two vision encoders, MedImageInsight (MI2) and RAD-DINO representing the two major encoder paradigms CLIP and DINOv2, on up to 3.5M chest x-rays from a single institution, holding compute and evaluation protocols constant. We evaluate on classification (radiology findings, lines and tubes),  segmentation (lines and tubes), and radiology report generation. While prior work has primarily focused on tasks related to radiology findings, we include lines and tubes tasks to counterbalance this bias and evaluate a model’s ability to extract features that preserve continuity along elongated structures. Our experiments show that MI2 scales more effectively for finding-related tasks, while RAD-DINO is stronger on tube-related tasks. Surprisingly, continually pretraining MI2 with both reports and structured labels using UniCL improves performance, underscoring the value of structured supervision at scale. We further show that for some tasks, as few as 30k in-domain samples are sufficient to surpass open-weights foundation models. These results highlight the utility of center-specific continual pretraining, enabling medical institutions to derive significant performance gains by utilizing in-domain data.
\end{abstract}
\section{Introduction}
Foundation models have shown strong potential in computer vision by leveraging large-scale pretraining for broad adaptability. Models trained on massive datasets like LAION-5B \citep{schuhmann_laion-5b_2022}, with billions of image–text pairs, achieve impressive zero-shot and few-shot performance on tasks such as ImageNet classification \citep{radford_learning_2021}. Two main pretraining paradigms dominate: image–text contrastive learning (e.g., CLIP \citep{radford_learning_2021}) and image-only self-supervised learning (e.g., DINOv2 \citep{oquab_dinov2_2024}). These differ in inputs, scalability, and downstream performance: CLIP excels at classification and retrieval, while DINOv2 performs better on segmentation and detection \citep{bolya_perception_2025, cherti_reproducible_2023, jiang_clip_2024, tong_eyes_2024}. In recent years, medical researchers have increasingly adopted foundation models to boost performance across diverse clinical tasks and address data scarcity and annotation challenges \citep{codella_medimageinsight_2024, perez-garcia_exploring_2025, lin_pmc-clip_2023, zhang_biomedclip_2025, zedda_radio_2025, moutakanni_advancing_2024}. Unlike general-domain datasets, chest X-ray (CXR) datasets typically contain only hundreds of thousands to a few million images, raising questions about how well general-domain pretraining insights transfer. We conduct a controlled comparison of two leading CXR pretraining approaches: MedImageInsight (MI2), which uses CLIP-style image–text contrastive learning \citep{codella_medimageinsight_2024}, and RAD-DINO, based on DINOv2-style image-only self-supervision \citep{perez-garcia_exploring_2025}. Both provide open weights and have shown state-of-the-art performance on public benchmarks. Our study leverages INST-CXR-BENCH, a large internal dataset of 4M de-identified CXR–report pairs, enabling control over data source and distribution, an advantage over prior work that compares models trained on heterogeneous datasets with varying compute budgets \citep{codella_medimageinsight_2024, perez-garcia_exploring_2025, huang_sc_shen_l_lungren_mp_yeung_s_gloria_2021, zhang_contrastive_2022, bannur_shruthi_learning_2023}. To ensure fairness, we use identical computational resources for both models. In our experiments, \Cref{sec:experiments}, we continually pretrain with both approaches on INST-CXR-BENCH and evaluate performance across multiple tasks and pretraining dataset sizes.

Our evaluation covers three task categories: classification, segmentation, and report generation. While prior work has focused mainly on radiological findings, we extend this by adding tasks on lines and tubes (l\&t) to probe learning of curve-continuity features, structures that preserve continuity along elongated objects (\Cref{sec:raddino}). We extract findings and l\&t labels from radiology reports using GPT. These labels primarily serve evaluation but also enable extending MI2 pretraining from CLIP to UniCL \citep{yang_unified_2022}, integrating structured labels with image–text contrastive learning. To support robust analysis, we construct a large test set of 400k samples from INST-CXR-BENCH, capturing long-tail findings and a diverse patient population. We establish scaling laws by analyzing performance across varying pretraining dataset sizes \citep{kaplan_scaling_2020}. For classification on INST-CXR-BENCH, as few as 30k samples of continual pretraining can surpass open-weight baselines. MI2 scales more effectively than RAD-DINO for findings classification, while both models show similar trends on l\&t classes. Notably, adding structured labels via UniCL significantly boosts MI2, an unexpected result given millions of image–report pairs. These patterns align with report generation experiments, \Cref{sec:maira_exps}, where we pair vision encoders with a Vicuna-13B LLaVA model \citep{liu_visual_2023, bannur_maira-2_2024, hyland_maira-1_2024}. Beyond our INST-CXR-BENCH dataset, we also evaluate the continually pretrained models on publicly available benchmarks. For findings classification, our updated models are on-par with or surpass the original open-weights models. For l\&t segmentation, RAD-DINO and both versions of MI2 outperform the original open-weights models.

Our findings reveal nuanced trade-offs between CLIP-style and DINOv2-style pretraining in medical imaging: 1. MI2 performs better on findings-related tasks; 2. RAD-DINO excels on l\&t; and 3. Adding label supervision via UniCL significantly improves MI2 performance on l\&t. More broadly, medical foundation models would benefit from training and evaluation on substantially larger and more diverse datasets than are common today. This need for scale is amplified by center-specific factors, including: (i) variability in image characteristics from scanners, protocols, and resolution; (ii) population-level differences such as age and ethnicity; and (iii) label distribution shifts, including rare conditions and reporting styles. In this context, continually pretraining center-specific foundation models on in-domain data, even with as few as 30k samples, can outperform open-weight models, underscoring current limitations in generalization of CXR foundation vision encoders.
\section{Related work}
Recent work has explored scaling laws for vision transformers (ViTs) \citep{zhai_scaling_2022} and self-supervised pretraining methods in general domains \citep{cherti_reproducible_2023}. For instance, \cite{fan_scaling_2025} show DINOv2 scales more favorably than CLIP with respect to both dataset size and model capacity at large scales involving billions of samples and parameters. However, these studies are primarily conducted at internet-scale datasets and with billion-parameter models, whereas medical imaging pretraining typically operates in a very different regime: model sizes of 0.3B parameters (i.e., ViT-L scale) or fewer and datasets that are several orders of magnitude smaller. To our knowledge, we are the first to systematically study the scaling behavior of pretraining vision encoders on CXR datasets up to millions of samples.

Existing work on scaling in medical imaging has largely focused on supervised learning. For example, \cite{cho_how_2016} studied the scaling of convolutional neural networks trained with supervised learning on limited medical data. \cite{xu_elixr_2023} as well as \cite{sellergren_simplified_2022} explore the scaling behavior of a linear findings classifier applied to a frozen vision encoder backbone, and also perform end-to-end fine-tuning with the same (unfrozen) encoder. In contrast, our work examines the effect of dataset size in the context of vision encoder pretraining using modern transformer-based architectures and a large-scale, single-modality medical image dataset.

Several recent efforts have implicitly compared the performance of DINOv2 and CLIP in medical domains. However, these comparisons are often confounded by differences in pretraining data and potentially compute budget. For instance, RAD-DINO and MI2 both evaluate multiple pretrained models, but the models are trained on different datasets, making direct performance comparisons difficult. Moreover, MI2 focuses on classification and retrieval tasks, which are known to favor CLIP-style contrastive learning approaches, potentially biasing conclusions. Our study addresses this by fairly comparing models trained with CLIP, DINOv2, and UniCL under controlled compute budgets, using consistent data sources, and a variety of tasks.

Last, the importance of considering layer-wise differences in representation quality has been highlighted in recent studies such as \cite{bolya_perception_2025}, which showed that different layers of a vision transformer can capture different types of features and exhibit variable downstream performance. However, this perspective has not been thoroughly explored in the medical domain. We are the first to incorporate this consideration into a systematic comparison of medical vision encoders, revealing insights that are potentially obscured when evaluating only the final layer representations.
\section{Method}
\label{sec:method}
\subsection{MedImageInsight}
\label{sec:m2i}
MedImageInsight (MI2) is a CLIP-style contrastive pretraining approach built on the Unified Contrastive Learning (UniCL \citep{yang_unified_2022}) framework. The open-weights MI2 vision encoder was trained on approximately 500k CXR image-text and image-label pairs plus 3.3M samples from various other medical imaging modalities. The open-weights model version of MI2 will be abbreviated with MI2 OWM throughout the paper. MI2 replaces the standard ViT backbone with a dual-attention ViT (DAViT) \citep{ding_davit_2022}, a hierarchical vision transformer that is claimed to be better suited for medical imaging tasks, particularly given the limited size of domain-specific datasets. CLIP-like models jointly train an vision encoder and a text encoder by projecting both modalities into a shared feature space. A contrastive loss, following the InfoNCE formulation \citep{oord_representation_2019}, aligns matched image–text pairs while pushing apart unmatched ones. UniCL extends the CLIP framework to support image–label contrastive learning. In MI2, structured categorical labels (e.g., disease annotations) are used as input to the text encoder, in the same way as radiology reports. The labels are represented by their category names or a list of names. These labels are tokenized and embedded by the text encoder, allowing the model to learn from both image–text and image–label pairs. Empirically, MI2 outperforms RAD-DINO and other CXR foundation vision encoders \citep{zhang_biomedclip_2025, moor_med-flamingo_2023} across a wide range of tasks including classification, retrieval, and findings generation, establishing it as the current state-of-the-art on finding-related tasks. This aligns with broader findings in the literature suggesting that CLIP-style models tend to excel at classification and retrieval, while DINOv2-style models may offer advantages in tasks that require dense outputs like segmentation \citep{jiang_clip_2024}. We used the open-weights MI2 weights as a starting point for continual pretraining as described here: \url{https://techcommunity. microsoft.com/blog/healthcareandlifesciencesblog/discovering-the-power-of-finetuning-medimageinsight-on-your-data/4395057}

\subsection{RAD-DINO}
\label{sec:raddino}
RAD-DINO \citep{perez-garcia_exploring_2025} is a self-supervised image-only pretraining approach for CXRs based on the DINOv2 \citep{oquab_dinov2_2024} approach. The open-weights RAD-DINO vision encoder was trained on $\sim$840k frontal and lateral CXRs with slight adjustments of the DINOv2 augmentations to be more suitable for CXRs. The open-weights model version of RAD-DINO will be abbreviated with RAD-DINO OWM throughout the paper. RAD-DINO inherits the core architectural and training principles of DINOv2, including self-distillation with ViT backbones \citep{caron_emerging_2021}, and masked image modeling in the style of iBOT \citep{zhou_ibot_2022}. There are two ViTs, the student and teacher networks. During training, multiple augmented views of each CXR are generated using radiology-specific transformations such as larger crop sizes and less severe blurring. There are three different parts of the loss function: (i) cross-entropy loss between the teacher’s and student’s CLS token, (ii) masked image loss where a subset of image patches is masked, and the student is trained to match the teacher’s representations of the masked tokens, (iii) the so-called KoLeo regularizer that encourages optimal use of the feature space. The teacher is updated via an exponential moving average of the student instead of gradient descent. RAD-DINO uses a ViT-Base model for both the student and teacher. At inference time, only the teacher network is used. At the time of its release, RAD-DINO outperformed both purely image-trained and image-text contrastive models \citep{bannur_shruthi_learning_2023, zhang_biomedclip_2025, tiu_expert-level_2022, zhou_advancing_2023} across a wide range of tasks, including findings classification, metadata classification, segmentation, and report generation. These results challenge the assumption that supervision via radiology reports is necessary for training high-performing vision encoders. The RAD-DINO checkpoint (including the DINO heads) is available at: \url{https://huggingface.co/microsoft/RAD-DINO}. We use the DINOv2 codebase for continual pretraining: \url{https://github.com/facebookresearch/dinov2}.

\begin{figure}[!b]
    \centering
    \begin{subfigure}[h]{0.19\textwidth}
        \includegraphics[height=3cm]{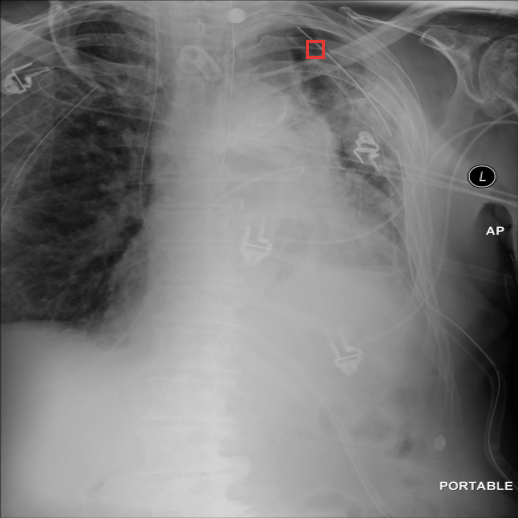}
        \caption{\scriptsize Original image}
    \end{subfigure}
    \hfill
    \begin{subfigure}[h]{0.19\textwidth}
        \includegraphics[height=3cm]{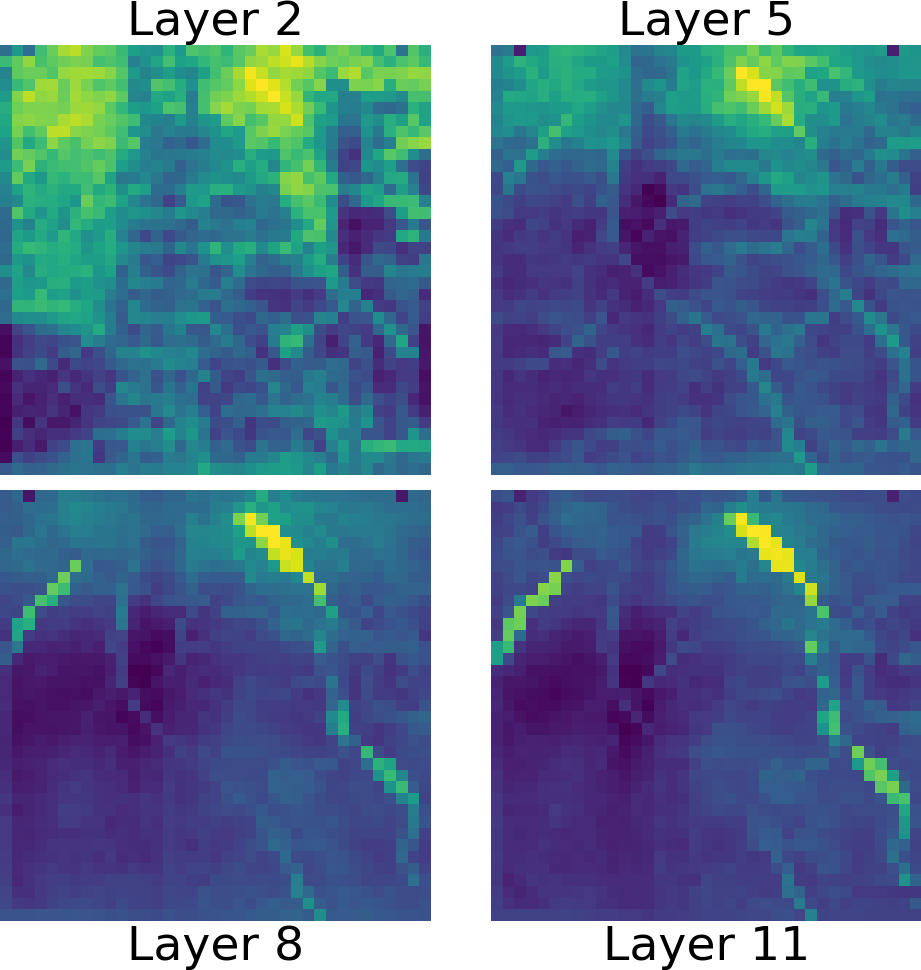}
        \caption{\scriptsize RAD-DINO 30k}
    \end{subfigure}
    \hfill
    \begin{subfigure}[h]{0.19\textwidth}
        \includegraphics[height=3cm]{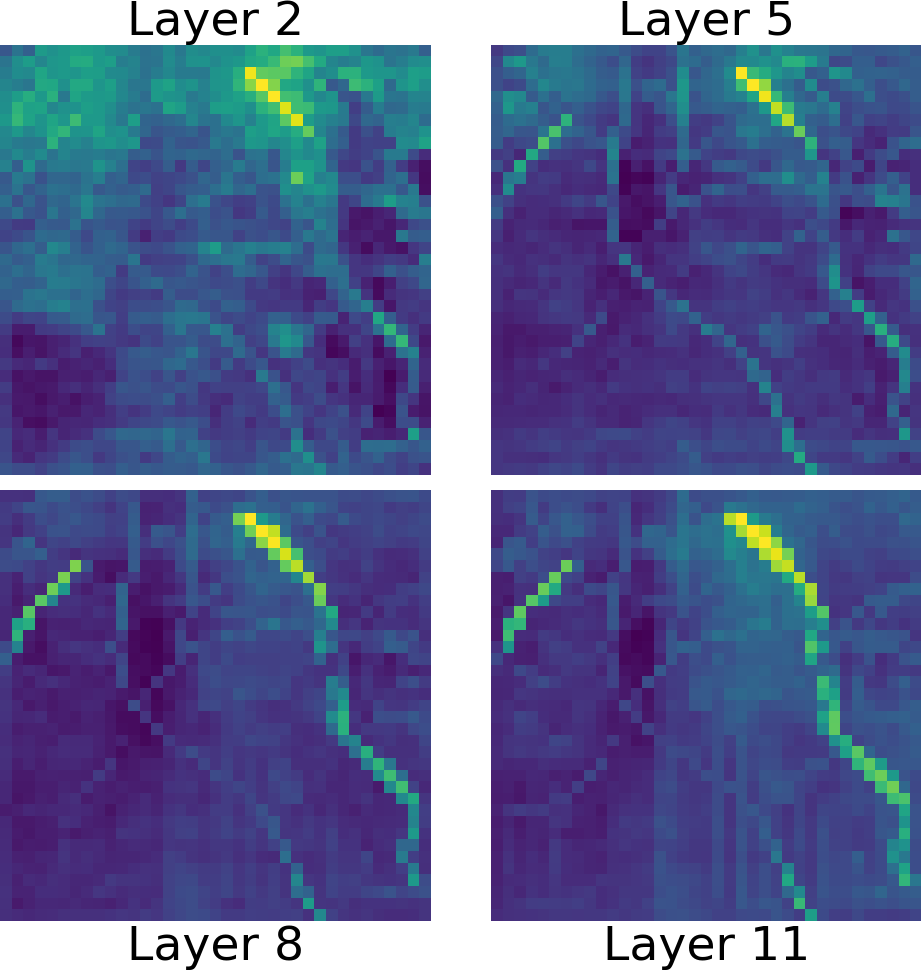}
        \caption{\scriptsize RAD-DINO 3.5M}
    \end{subfigure}
    \hfill
    \begin{subfigure}[h]{0.19\textwidth}
        \includegraphics[height=3cm]{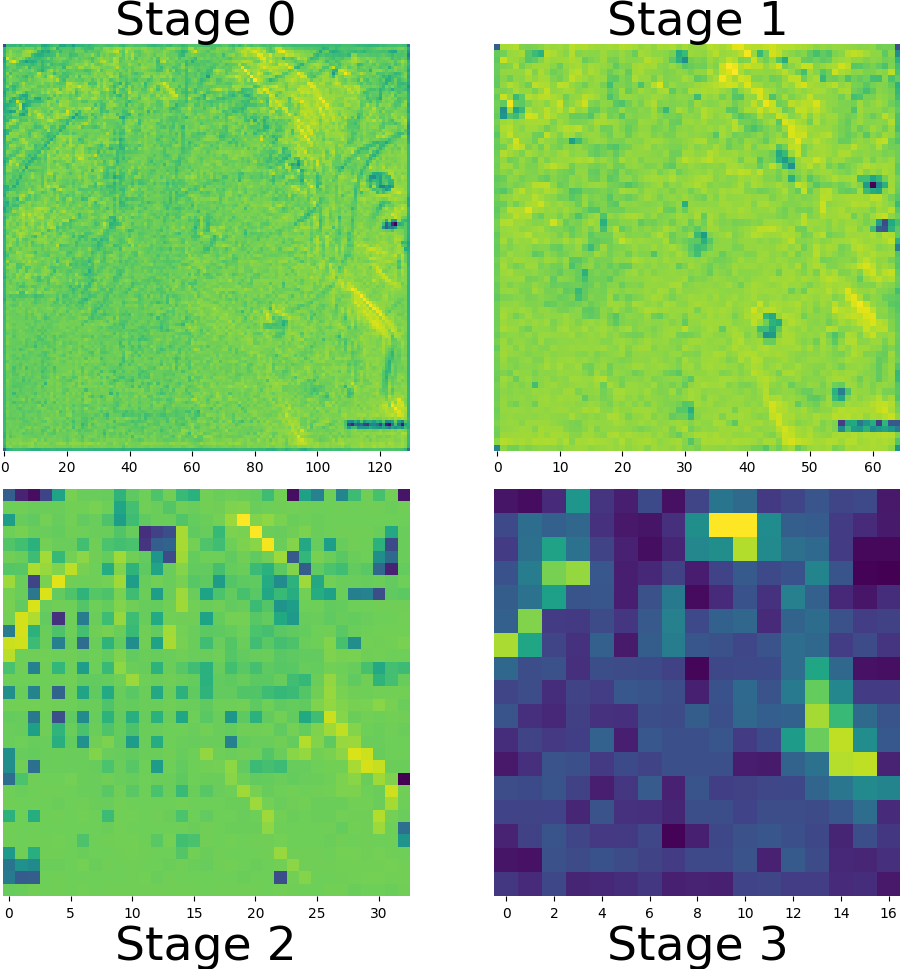}
        \caption{\scriptsize MI2 reports 30k}
    \end{subfigure}
    \hfill
    \begin{subfigure}[h]{0.19\textwidth}
        \includegraphics[height=3cm]{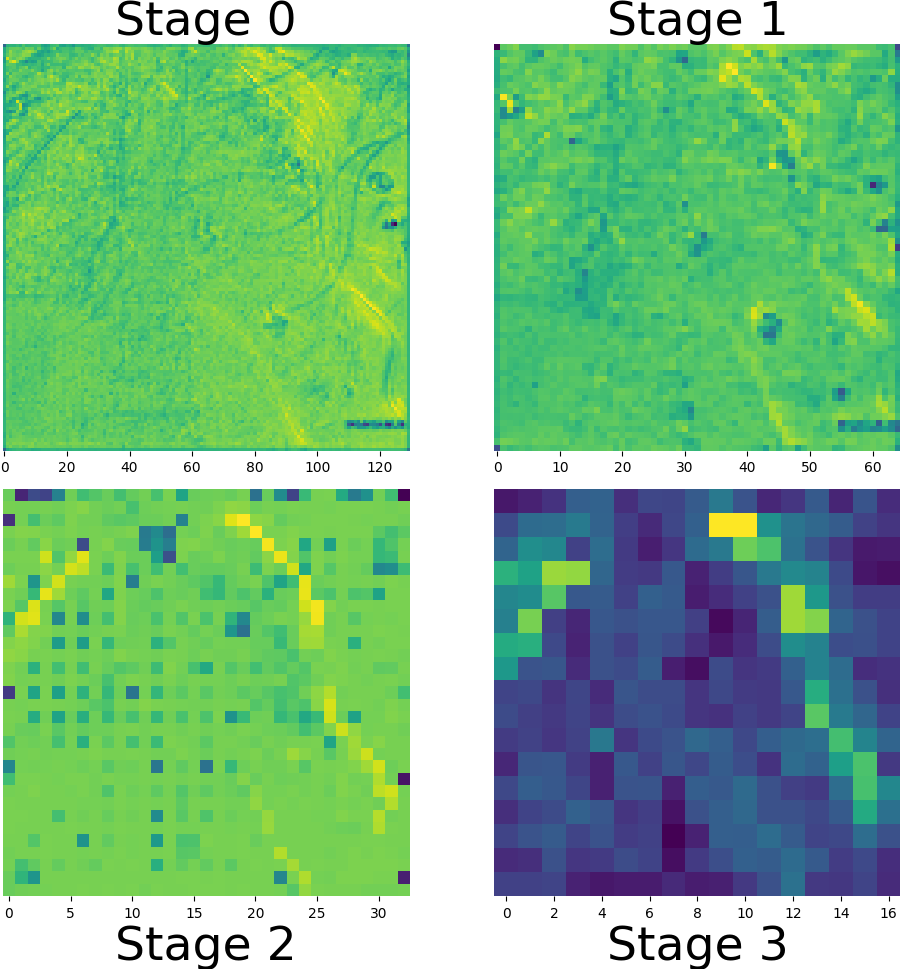}
        \caption{\scriptsize MI2 reports 3.5M}
    \end{subfigure}
    \caption{We visualize the cosine similarity maps between the patch marked with a red box (tip of a chest tube) and all other patches extracted for four layers from RAD-DINO and MI2, each pretrained with either 30k or 3.5M images. We argue that an ideal feature map should: (i) highlight all patches belonging to the chest tube on the right side of the image, (ii) highlight all the patches belonging to the chest tube on the left side of the image, (iii) should not highlight any other tubes or structures. For both models pretrained with 3.5M the last feature maps are the closest to the ideal feature map described above, all other feature maps seem to highlight additional tubes and structures. While more prevalent in MI2, at 30K both models miss patches associated with chest tubes, underscoring the benefits of large pretraining datasets.}
    \label{fig:dino_clip_comparison}
\end{figure}

Throughout \Cref{sec:experiments}, RAD-DINO demonstrates strong performance on tasks involving lines and tubes (l\&t), performing on par with MI2 even in l\&t classification. This is especially notable given the well-established advantage of CLIP-pretrained models over DINOv2 on classification tasks \citep{bolya_perception_2025}. We hypothesize that RAD-DINO benefits from self-distillation with masked multi-view objectives, which encourages the learning of curve-continuity features (CCF), features that preserve continuity along elongated structures such as l\&t, see \Cref{fig:dino_clip_comparison}. These features are particularly well-suited for tasks like tube tip localization and segmentation, where even small discontinuities can result in significant penalties. In contrast, several aspects of CLIP may inhibit the learning of CCF. First, CLIP aligns global features from the image and text encoders, which may fail to capture fine-grained structural details \citep{huang_sc_shen_l_lungren_mp_yeung_s_gloria_2021}. Second, chest X-ray reports often omit or only sparsely mention medical devices, frequently lacking the detailed descriptions needed for robust alignment.

\subsection{Continual pretraining and scaling laws}
\label{sec:scaling_laws}
Previous work has shown that the performance of large models improves predictably with scale, following power-law relationships with respect to model size, dataset size, and compute budget \citep{kaplan_scaling_2020}. This has enabled performance extrapolation from early training curves, providing a framework to guide the development of increasingly capable models. In particular, dataset size has been identified as a dominant factor in scaling performance. \cite{bansal_data_2022} argue that dataset size contributes more significantly than architecture or model size in the domain of neural machine translation. Similar findings in the vision domain confirm the critical role of data quantity in driving performance gains \citep{zhai_scaling_2022}. In this work, we focus specifically on dataset size scaling laws while keeping compute and model size fixed. This decision is motivated by two key factors: (i) we use pretrained RAD-DINO and MI2 checkpoints as our starting point, which constrains our ability to vary model size; and (ii) prior work consistently demonstrates that data quantity is the most influential factor in driving performance improvements. When applicable (e.g., \Cref{sec:finginds_internal}), we fit a power-law of the form $f(x) = \alpha x^k$, where $f(x)$ is a performance metric (e.g., AUPRC) obtained by evaluating a frozen encoder on a downstream task, pretrained on a dataset of size $x$. However, recent studies caution against overgeneralizing the predictive power of scaling laws. For instance, \cite{caballero_broken_2023} and \cite{alabdulmohsin_revisiting_2022} show that power-law behavior is often confined to a narrow region of the parameter space, with saturation effects becoming apparent at larger scales. Similarly, \cite{lourie_scaling_2025} demonstrate that downstream performance may exhibit emergent behavior, saturation, or even inverse scaling, where increased scale degrades performance potentially due to a distribution shift and catastrophic forgetting.

\section{INST-CXR-BENCH dataset creation}
For pretraining and evaluation we use a large internal dataset consists of 3.1M CXR studies sourced from the Mayo Clinic, with approximately 23\% of the studies containing at least one line or tube. The data was split on a patient level into 80\% for training, 10\% for validation, and 10\% for testing. Each study contains longitudinal information, incorporating current frontal and lateral images, prior frontal images, prior reports, and clinical context such as indication and comparison sections. From the 3.1M studies we create a dataset INST-CXR-BENCH containing approximately 4 million CXR images and associated reports, where the same patient can contribute multiple images and reports. The images are divided into frontal (62\%) and lateral (38\%) images. Patient sex is divided into three categories: 'Male' (49\%), 'Female' (48\%), and 'Other' (2\%). Patient ethnicity consists of seven categories: 'White' (87\%), 'Black or African American' (3\%), 'Asian' (1\%), 'Asian - Far East' (1\%), 'Native American and Pacific Islander' (1\%), and 'Asian - Indian Subcontinent' ($<$1\%). Patient age has the following distribution: ‘$<$20' (2\%), '20-30' (7\%), '30-40' (9\%), '40-50' (13\%), ‘50-60’ (22\%), '60-70' (22\%), '70-80' (17\%), '80-89' (7\%), '89+' (1\%). The images and reports were created between 2013 and 2023: '2007-2012' (35\%), '2013-2017' (25\%), '2018-2023' (37\%). There are 16 different departments that ordered CXRs, the major six are 'Internal Medicine' (14\%), 'Emergency Medicine' (13\%), 'Cardiovascular Diseases' (9\%), 'Radiology' (9\%), 'Family Medicine' (8\%), 'General Practice' (6\%).  The patients are grouped into ‘inpatient’ (58\%) and ‘outpatient ‘(40\%). The images were acquired by scanner from 19 different manufacturers, the major seven are: 'FUJIFILM Corporation' (38\%), 'Carestream Health' (24\%), 'GE Healthcare' (15\%), 'SIEMENS' (8\%), 'Philips' (6\%), 'Canon Inc.' (3\%). CXR DICOM images were converted to PNG and resized to 518px using B-spline interpolation with antialiasing. To ensure patient privacy, white boxes were overlaid on the images to mask identifying information such as text, facial features, or other visual elements that could potentially reveal a patient’s identity. Intensities were normalized to an 8-bit range. GPT-4o \citep{openai_gpt-4o_2024} was used to parse and clean the reports into structured JSON, handling inconsistent formatting, duplications, and artifacts of the EHR storing process. Each frontal image was linked to corresponding lateral and prior images, when available. A de-duplication step retained one image per type (frontal, lateral, prior frontal) per visit, prioritizing original images with complete metadata. Last, Fastdup$^1$\footnotetext{1 \url{https://www.visual-layer.com/}} was used to detect and remove $\sim$6\% of outlier images, such as blank or non-chest X-rays.

\section{Experiments}
\label{sec:experiments}
To better understand how foundation models in medical imaging scale with data, we perform continual pretraining starting from two strong open-weights baselines: MI2 \citep{codella_medimageinsight_2024} and RAD-DINO \citep{perez-garcia_exploring_2025}. We progressively pretrain both models on five strictly nested subsets of INST-CXR-BENCH: 30k, 50k, 100k, 1M, and 3.5M image-report pairs. Each model is trained for equal wall-clock time on four nodes, with eight H100 GPUs per node, with a batch size of 1280 (40 samples per GPU). The training durations for each data size are: 0.33, 0.58, 1.17, 11.67, 40.83 hours, respectively, with a standard deviation of 8\% in training time. These durations correspond to 15 epochs of MI2 training, and RAD-DINO is trained for an equivalent amount of time by adjusting its epoch count accordingly. For MI2, we pretrain two variants: Image-report contrastive learning (standard CLIP) and image-report plus image-label following the UniCL approach \citep{yang_unified_2022}, where each image is seen once paired with its report and a second time paired with a label per epoch, if available. We adjust the number of epochs to ensure equal wall-clock time. As a case study, we use tube presence labels (e.g., “Nasogastric Tube, Endotracheal Tube”) extracted via GPT (\Cref{sec:gpt_extraction}); 23\% of CXRs include at least one line or tube. All MI2 text inputs include a view-position prefix. For 30k, 50k, and 100k, we create three random dataset subsets resulting in three encoders per pretraining approach. For 1M and 3.5M, we train one encoder each. Results are compared against the original open-weight models (OWM) (Figures \ref{fig:mayo_findings}–\ref{fig:mayo_findings_generation}). Throughout the following section, we clearly distinguish which results are statistically significant, and whenever we label a result as significant, the claim is supported by the significance test described in \Cref{sec:signify_test}.

\subsection{Classification}
\label{sec:classification}
To compare embedding quality, we run classification experiments using frozen backbones, isolating feature quality without fine-tuning. Our goal is to evaluate MI2 and RAD-DINO under identical downstream conditions.
We avoid using the [CLS] token, as prior work \citep{bolya_perception_2025} shows different layers encode different information. Instead, we extract features from four layers per encoder. MI2, based on DAViT, outputs multi-scale features, so we apply convolutions and linear projections to unify dimensions to 33×33×1024 (third block size). While RAD-DINO uses a ViT-B backbone where all feature maps have identical shapes, we still apply the same projection strategy for fairness, using layers 2, 5, 8, and 11 (zero-indexed). Details are in \Cref{tab:comparison}.
Projected features from the four blocks are concatenated (4096 dims for MI2, 3072 for RAD-DINO), then reduced via attention pooling to a single token per task, followed by a linear classifier. We train for 25 epochs with batch size 512. Each experiment is repeated three times: for 30k, 50k, and 100k using different encoders and seeds; for 1M and 3.5M using the same encoder with three seeds.

\subsubsection{Findings classification on INST-CXR-BENCH-FIND-CLASS}
\label{sec:finginds_internal}
We use a 2M subset of frontal CXRs from INST-CXR-BENCH (75\%/25\% train/test images) for training and evaluating findings classification models. Subsequently, we will call the subset INST-CXR-BENCH-FIND-CLASS.  We choose 19 findings categories covering a wide range of appearances (some are more diffuse/texture like, others are more localized/shape like) and areas of a CXR (from the esophagus to the diaphragm). Each finding has at least 10k examples in the train set. All labels are extracted from the paired reports as described in \Cref{sec:gpt_extraction}. 
\Cref{tab:mayo_findings_class} contains the list of all findings we consider and their prevalences. Note: While we are varying the number of pretraining samples, the amount of samples to train the classification model is always the same.

\begin{figure}[h]
    \centering
    \begin{minipage}[h]{0.49\linewidth}
        \centering
        \includegraphics[width=\linewidth]{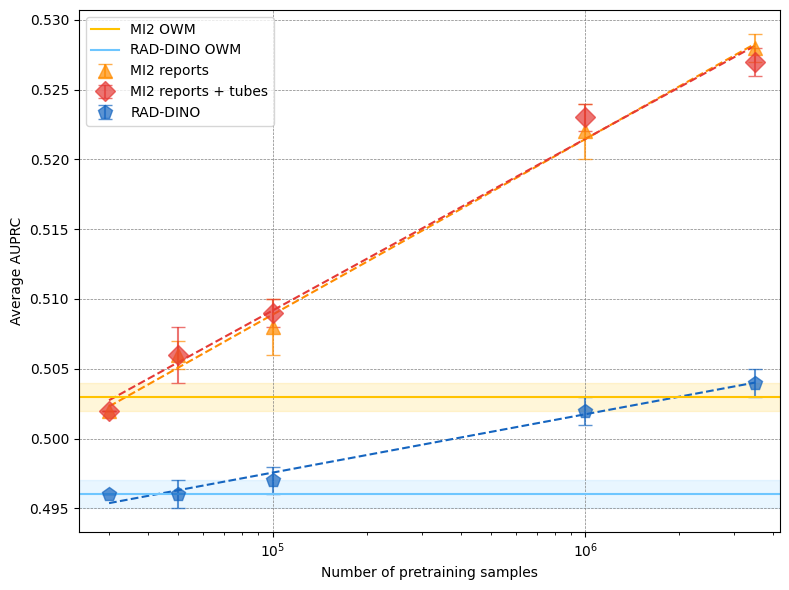}
    \end{minipage}
    \hfill
    \begin{minipage}[h]{0.49\linewidth}
        \centering
        \includegraphics[width=\linewidth]{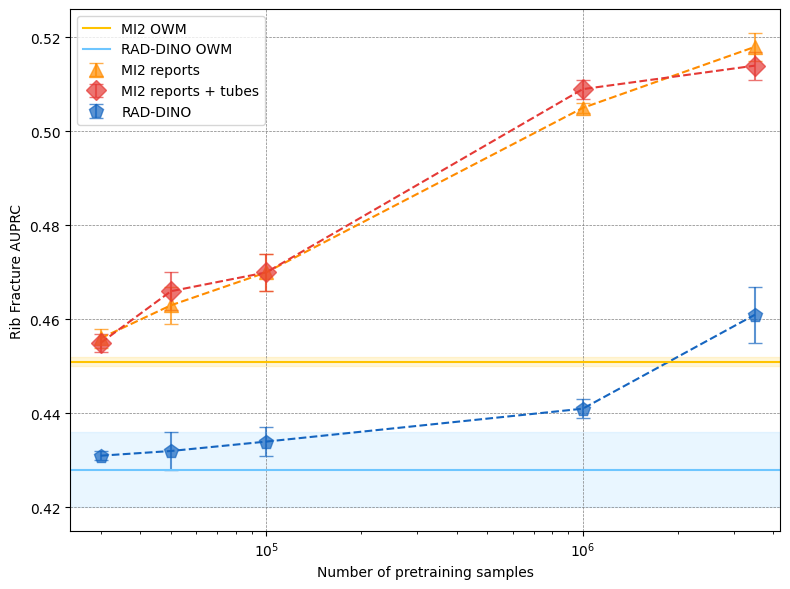}
    \end{minipage}
    \caption{Findings classification performance on INST-CXR-BENCH-FIND-CLASS as a function of vision encoder pretraining with increasing sample sizes from INST-CXR-BENCH. Left: AUPRC averaged across 20 finding tasks. Both MI2 models have slope $k$=0.012 and intercept $\alpha$=0.447, RAD-DINO has slope $k$=0.004 and intercept $\alpha$=0.466, i.e., MI2 scales about three times better than RAD-DINO. Right: AUPRC for finding rib fracture (prevalance 2.3\%), which shows the greatest improvement when pretrained with 3.5M samples.}
    \label{fig:mayo_findings}
\end{figure}

In \Cref{fig:mayo_findings} (left), we observe clear power-law behavior when plotting classification performance against dataset size on a log scale. The scaling trends appear linear, indicating predictable and consistent gains as more data is used for continual pretraining. In agreement with \citep{bolya_perception_2025}, MI2 significantly outperforms RAD-DINO on this classification task starting at the pretraining dataset size of 100k. Different MI2 variants (CLIP vs UniCL) perform comparably, suggesting that the additional supervision using tube presence labels has no negative effect on findings classification. Comparing the power law fits in \Cref{fig:mayo_findings} (left) we find that MI2 scales about three times better than RAD-DINO. In addition, we find that already with 100k images, continual pretraining can significantly outperform public foundation model checkpoints, highlighting the value of domain-specific adaptation. While average AUPRC improvements may appear modest, class-specific gains can be substantial. In \Cref{fig:mayo_findings} (right), we show scaling laws for the binary task of rib fracture classification. MI2 continually pretrained with 3.5M images shows an improvement of 6\% compared to MI2 open-weights model, also see \Cref{tab:mayo_findings_class}. For MI2, 30k samples of INST-CXR-BENCH are sufficient to significantly outperform the open-weights model. In contrast, for RAD-DINO, 100k samples are needed to outperform the open-weights model. In \Cref{tab:mayo_findings_class}, we compare models trained with all of INST-CXR-BENCH's data (3.5M samples) vs the open-weights models. We find that for all 20 binary tasks both variants of MI2 pretrained on INST-CXR-BENCH outperform the open-weights model of MI2 as well as RAD-DINO. The most noticeable improvements (greater than 5\%) are observed in the binary classification tasks of detecting pneumothorax, enlarged pulmonary artery, and rib fracture. For findings classification (see also \Cref{sec:findings_vindr}), we identify several classes with low AUPRC, likely due to noisy labels. We attribute this noise to three main sources: (i) inter-reader variation among radiologists, (ii) inaccurate original reports, and (iii) errors introduced during the GPT-based extraction of structured labels (see \Cref{sec:gpt_extraction}). In \Cref{sec:metadata}, we report all results from section stratified by various metadata variables.
\subsubsection{Tube presence classification on INST-CXR-BENCH-TUBE-CLASS}
\label{sec:tubes_internal}
We use a 1M subset of frontal CXRs from INST-CXR-BENCH (90\%/10\% train/test images) for training and evaluating tube presence classification models. Subsequently, we will call this subset INST-CXR-BENCH-TUBE-CLASS. L\&t prevalences are ranging from 0.56\% to 14.03\%, see \Cref{tab:mayo_lines_and_tubes_class} for exact numbers. All labels are extracted from the paired reports as described in \Cref{sec:gpt_extraction}. Note: While we are varying the number of pretraining samples, the amount of samples to train the classification model is always the same.

\begin{figure}[h]
    \centering
    \begin{minipage}[h]{0.49\linewidth}
        \centering
        \includegraphics[width=\linewidth]{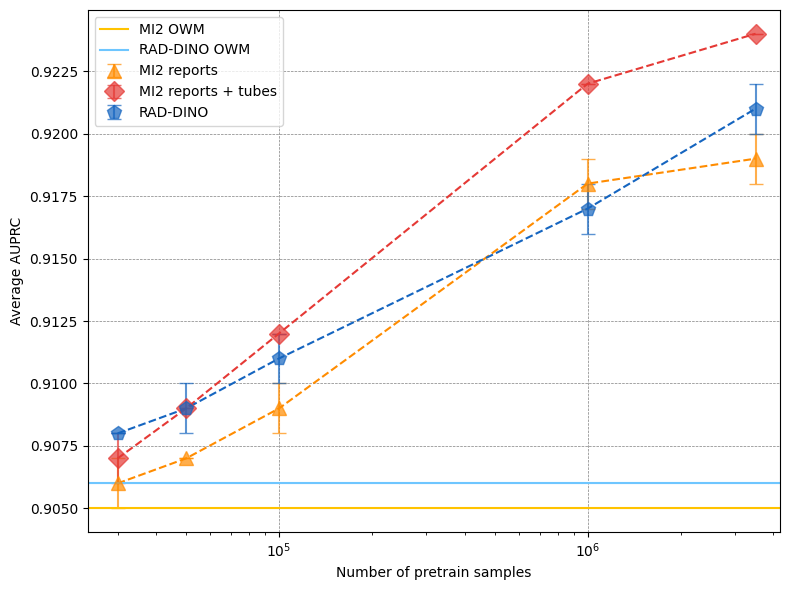}
    \end{minipage}
    \hfill
    \begin{minipage}[h]{0.49\linewidth}
        \centering
        \includegraphics[width=\linewidth]{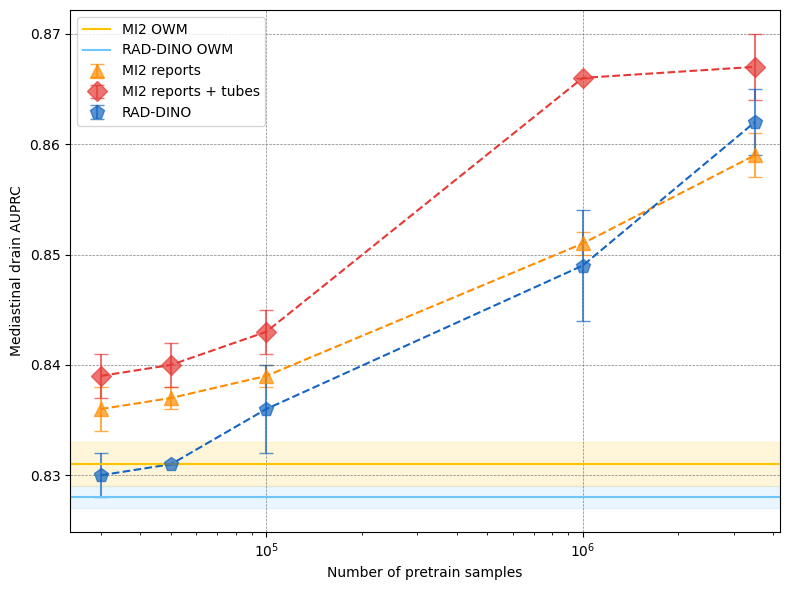}
    \end{minipage}
    \caption{Tube presence classification on INST-CXR-BENCH-TUBE-CLASS as a function of vision encoder pretraining with increasing samples sizes from INST-CXR-BENCH. Left: AUPRC averaged across 11 lines and tubes tasks. Right: AUPRC for the tube mediastinal drain (Prevalance 5.6\%). Task with the highest improvement when pretrained with 3.5M samples.}
    \label{fig:mayo_tubes}
\end{figure}

RAD-DINO consistently has a higher or on par average AUPRC compared to MI2 trained solely on reports. Only at 1M, the performance is comparable, see \Cref{fig:mayo_tubes}. Overall, MI2 seems to saturate faster with increasing pretraining data than RAD-DINO, a similar observation was made in \cite{fan_scaling_2025}. Furthermore, RAD-DINO and MI2 trained on INST-CXR-BENCH begin to significantly outperform open-weights models at 100k pretraining samples. Adding tube presence labels to report-based MI2 improves the performance of MI2, surpassing RAD-DINO on average AUPRC. MI2 trained with both reports and tube presence labels outperforms the open-weights MI2 model at around 50k samples, significantly earlier than MI2 trained only with reports. It is important to note that for the average AUPRC none of the three pretraining methods significantly outperforms the other two. While the average AUPRC gains reported in \Cref{fig:mayo_tubes} (left) are modest, we observe more substantial improvements for less prevalent and harder-to-detect tube types, such as intra-aortic balloon pumps (small structure) and mediastinal drains (often obscured by the spine), see \Cref{fig:mayo_tubes} (right) and \Cref{tab:mayo_lines_and_tubes_class}. For the mediastinal drain in particular (\Cref{fig:mayo_tubes} right), MI2 trained with both reports and tube presence labels significantly outperforms RAD-DINO starting at a pretraining dataset size of 30k and MI2 trained without tube presence labels starting at a pretraining dataset size of 100k, highlighting the benefit of incorporating GPT-extracted labels during pretraining. In \Cref{tab:mayo_lines_and_tubes_class} we compare models trained with all of INST-CXR-BENCH (3.5M samples) vs the open-weights models. We find that for all 11 binary l\&t detection tasks MI2 trained with reports and tube presence labels is outperforming all other models or is on par with RAD-DINO.
\begin{table}[h]
\centering
\small
\caption{Findings classification on the INST-CXR-BENCH-FIND-CLASS subset. Comparison of the open-weights models (OWM) and encoders pretrained with the full INST-CXR-BENCH dataset (3.5M samples).}
\label{tab:mayo_findings_class}
\begin{tabular}{llllllll}
\toprule
Finding & HI & ILD & ATL & CAB & PE & PTX & AD \\
Prevalence & 1.57\% & 1.97\% & 14.34\% & 1.03\% & 10.39\% & 2.61\% & 0.35\% \\
\midrule
MI2 OWM & 35.8\textsubscript{0.5} & 21.2\textsubscript{0.3} & 73.6\textsubscript{0.1} & 16.2\textsubscript{0.1} & 84.4\textsubscript{0.1} & 76.8\textsubscript{0.1} & 16.3\textsubscript{0.3} \\
MI2 reports & \textbf{37.6\textsubscript{0.2}} & \textbf{23.6\textsubscript{0.4}} & \textbf{74.9\textsubscript{0.0}} & \textbf{18.3\textsubscript{0.3}} & \textbf{85.4\textsubscript{0.0}} & \textbf{81.6\textsubscript{0.1}} & \textbf{20.2\textsubscript{0.1}} \\
MI2 reports + tubes & \textbf{37.3\textsubscript{0.2}} & 23.3\textsubscript{0.3} & \textbf{74.9\textsubscript{0.0}} & \textbf{18.0\textsubscript{0.2}} & 85.3\textsubscript{0.0} & \textbf{81.6\textsubscript{0.2}} & 18.9\textsubscript{0.5} \\
RAD-DINO OWM & 35.5\textsubscript{0.4} & 21.1\textsubscript{0.2} & 73.0\textsubscript{0.2} & 15.1\textsubscript{0.3} & 84.3\textsubscript{0.2} & 74.9\textsubscript{0.3} & 14.8\textsubscript{1.1} \\
RAD-DINO & 36.0\textsubscript{0.3} & 22.0\textsubscript{0.2} & 73.4\textsubscript{0.1} & 15.7\textsubscript{0.4} & 84.6\textsubscript{0.1} & 77.8\textsubscript{0.2} & 14.1\textsubscript{0.2} \\
\bottomrule
\end{tabular}

\begin{tabular}{llllllll}
\toprule
Finding & EPA & AC & OA & RF & BWT & HRN & SA \\
Prevalence & 0.61\% & 7.60\% & 8.83\% & 2.28\% & 0.60\% & 0.81\% & 1.18\% \\
\midrule
MI2 OWM & 28.3\textsubscript{0.4} & 58.4\textsubscript{0.3} & 57.0\textsubscript{0.2} & 45.1\textsubscript{0.1} & 10.3\textsubscript{0.1} & 65.1\textsubscript{0.3} & 78.4\textsubscript{0.1} \\
MI2 reports & \textbf{32.1\textsubscript{0.9}} & \textbf{61.4\textsubscript{0.0}} & \textbf{59.7\textsubscript{0.0}} & \textbf{51.8\textsubscript{0.3}} & {12.6\textsubscript{0.3}} & 68.6\textsubscript{0.5} & \textbf{80.5\textsubscript{0.1}} \\
MI2 reports + tubes & \textbf{32.5\textsubscript{0.7}} & \textbf{61.3\textsubscript{0.2}} & \textbf{59.6\textsubscript{0.1}} & 51.4\textsubscript{0.3} & \textbf{13.0\textsubscript{0.2}} & \textbf{69.3\textsubscript{0.3}} & \textbf{80.8\textsubscript{0.0}} \\
RAD-DINO OWM & 28.1\textsubscript{1.2} & 57.0\textsubscript{0.1} & 56.5\textsubscript{0.1} & 42.8\textsubscript{0.8} & 9.6\textsubscript{0.3} & 65.7\textsubscript{0.2} & 77.9\textsubscript{0.1} \\
RAD-DINO & 30.2\textsubscript{0.6} & 58.4\textsubscript{0.2} & 57.2\textsubscript{0.2} & 46.1\textsubscript{0.6} & 9.7\textsubscript{0.2} & 68.3\textsubscript{0.2} & 79.5\textsubscript{0.3} \\
\bottomrule
\end{tabular}

\begin{tabular}{llllllll}
\toprule
Finding & OP & VC & CM & DE & PDE & NF & AVG \\
Prevalence & 12.85\% & 2.71\% & 10.23\% & 2.10\% & 2.64\% & 7.30\% \\
\midrule
MI2 OWM & 75.9\textsubscript{0.0} & 26.1\textsubscript{0.1} & 76.3\textsubscript{0.1} & 42.3\textsubscript{0.2} & 36.5\textsubscript{0.2} & 81.3\textsubscript{0.1} & 50.3\textsubscript{0.1} \\
MI2 reports & \textbf{77.2\textsubscript{0.0}} & \textbf{27.5\textsubscript{0.5}} & \textbf{77.6\textsubscript{0.0}} & \textbf{45.0\textsubscript{0.2}} & \textbf{38.0\textsubscript{0.2}} & \textbf{82.9\textsubscript{0.0}} & \textbf{52.8\textsubscript{0.1}} \\
MI2 reports + tubes & 77.1\textsubscript{0.0} & \textbf{27.5\textsubscript{0.3}} & 77.4\textsubscript{0.2} & \textbf{45.0\textsubscript{0.3}} & \textbf{37.7\textsubscript{0.1}} & \textbf{82.8\textsubscript{0.1}} & \textbf{52.7\textsubscript{0.1}} \\
RAD-DINO OWM & 75.2\textsubscript{0.1} & 25.1\textsubscript{0.2} & 76.6\textsubscript{0.1} & 41.7\textsubscript{0.4} & 36.4\textsubscript{0.2} & 80.6\textsubscript{0.0} & 49.6\textsubscript{0.1} \\
RAD-DINO & 75.8\textsubscript{0.0} & 25.5\textsubscript{0.4} & 76.7\textsubscript{0.1} & 40.1\textsubscript{0.4} & 36.6\textsubscript{0.3} & 81.3\textsubscript{0.1} & 50.4\textsubscript{0.1} \\
\bottomrule
\end{tabular}

\small
HI: Hyperinflation, ILD: Interstitial Lung Disease Pattern, ATL: Atelectasis, CAB: Costophrenic Angle Blunting, PE: Pleural Effusion, PTX: Pneumothorax, AD: Adenopathy, EPA: Enlarged Pulmonary Artery, AC: Arterial Calcification, OA: Osseous Abnormalities, RF: Rib Fracture, BWT: Bronchial Wall Thickening, HRN: Hernia, SA: Subcutaneous Air/Emphysema, OP: Opacity, VC: Vascular Congestion, CM: Cardiomegaly, DE: Diaphragm Elevation, PDE: Pulmonary Edema, NF: No Finding, AVG: Mean of micro averaged AUPRC across 3 seeds.
\end{table}
\begin{table}[h]
\centering
\small
\caption{Tube presence classification on the INST-CXR-BENCH-TUBE-CLASS subset. Comparison of the open-weights models and encoders pretrained with the full INST-CXR-BENCH dataset (3.5M samples).}
\label{tab:mayo_lines_and_tubes_class}

\begin{tabular}{llllllll}
\toprule
Tube category & ETT & TT & NGT & SGC & CT & MD & IABP \\
Prevalence & 11.39\% & 3.23\% & 12.73\% & 5.17\% & 14.03\% & 5.56\% & 0.56\% \\
\midrule
MI2 OWM & 95.6\textsubscript{0.1} & 94.8\textsubscript{0.1} & 92.4\textsubscript{0.0} & 94.8\textsubscript{0.2} & 96.4\textsubscript{0.1} & 83.1\textsubscript{0.2} & 81.5\textsubscript{0.4} \\
MI2 reports & 96.2\textsubscript{0.1} & \textbf{95.5\textsubscript{0.0}} & \textbf{93.3\textsubscript{0.1}} & \textbf{95.5\textsubscript{0.1}} & 97.2\textsubscript{0.0} & 85.9\textsubscript{0.2} & \textbf{83.6\textsubscript{0.5}} \\
MI2 reports + tubes & \textbf{96.3}\textsubscript{0.1} & \textbf{95.7}\textsubscript{0.1} & \textbf{93.4}\textsubscript{0.0} & \textbf{95.7}\textsubscript{0.1} & \textbf{97.5}\textsubscript{0.0} & \textbf{86.7}\textsubscript{0.3} & \textbf{84.5}\textsubscript{0.3} \\
RAD-DINO OWM & 95.1\textsubscript{0.0} & 94.4\textsubscript{0.2} & 92.2\textsubscript{0.0} & 94.6\textsubscript{0.1} & 96.1\textsubscript{0.0} & 82.8\textsubscript{0.1} & 82.3\textsubscript{0.4} \\
RAD-DINO & 95.9\textsubscript{0.0} & \textbf{95.2\textsubscript{0.1}} & \textbf{93.3\textsubscript{0.1}} & \textbf{95.3\textsubscript{0.1}} & 97.0\textsubscript{0.0} & 86.2\textsubscript{0.3} & \textbf{85.6\textsubscript{0.5}}\\
\bottomrule
\end{tabular}

\begin{tabular}{lllllll}
\toprule
Tube category & IJ & PICC & SC & NT & AVG \\
Prevalence & 12.38\% & 13.49\% & 5.49\% & 14.04\% \\
\midrule
MI2 OWM & 81.8\textsubscript{0.2} & 95.9\textsubscript{0.1} & 81.7\textsubscript{0.7} & 97.6\textsubscript{0.0} & 90.5\textsubscript{0.0} \\
MI2 reports & 83.4\textsubscript{0.2} & \textbf{96.7\textsubscript{0.0}} & 85.6\textsubscript{0.2} & 98.0\textsubscript{0.0} & \textbf{91.9\textsubscript{0.1}} \\
MI2 reports + tubes & \textbf{84.3}\textsubscript{0.1} & \textbf{96.9}\textsubscript{0.0} & \textbf{87.0}\textsubscript{0.1} & \textbf{98.1}\textsubscript{0.0} & \textbf{92.4}\textsubscript{0.0} \\
RAD-DINO OWM & 82.0\textsubscript{0.1} & 96.1\textsubscript{0.0} & 83.5\textsubscript{0.3} & 97.6\textsubscript{0.0} & 90.6\textsubscript{0.0} \\
RAD-DINO & 83.8\textsubscript{0.1} & \textbf{96.9\textsubscript{0.0}} & 85.9\textsubscript{0.2} & 98.0\textsubscript{0.0} & \textbf{92.1\textsubscript{0.1}}\\
\bottomrule
\end{tabular}

\centering
\small
ETT: Endotracheal Tube, TT: Tracheostomy Tube, NGT: Nasogastric Tube, SGC: Swan-Ganz Catheter, CT: Chest Tube, \\
MD: Mediastinal Drain, IABP: Intra-Aortic Balloon Pump, IJ: Internal Jugular CVC, PICC: Peripherally Inserted Central Catheter, \\
SC: Subclavian CVC / Port-a-Cath, NT: No Tubes, AVG: Mean of micro averaged AUPRC across 3 seeds.
\end{table}
\subsubsection{Findings classification on holdout dataset VinDR}
\label{sec:findings_vindr}
In addition to the in-domain classification experiments in \Cref{sec:finginds_internal} and \Cref{sec:tubes_internal}, we evaluate finding classification performance on a public holdout dataset, called VinDR \citep{nguyen_vindr-cxr_2022}, comprising 9k CXR images (90\%/10\% train/test). Note: While we are varying the number of pretraining samples, the amount of samples to train the classification model is always the same. Due to the small size of the dataset and its potential domain mismatch with our training data, clear performance scaling trends are difficult to establish, see \Cref{sec:scaling_laws}. This reflects a broader challenge in medical imaging: small public benchmarks often provide limited insight into the generalization capabilities of foundation models. In \Cref{fig:vindr_findings_and_ranzcr_hausdorff} (left), RAD-DINO exhibits a reversed scaling trend. As discussed in \Cref{sec:scaling_laws}, this can occur when the target data distribution aligns more closely with that of a pretrained model. In this case, the open-weights RAD-DINO encoder appears better suited to the VinDR dataset, likely due to favorable pretraining data distribution and checkpoint selection. At larger scales, we observe signs of catastrophic forgetting, further suggesting a distribution shift. Both versions of MI2 are more in line with the expected trend: more pretraining data leads to better models. In contrast to the experiments in \Cref{sec:finginds_internal}, we need substantially more samples ($>$100k) to outperform the open-weights models. We also observe a notable effect when tube presence labels are included alongside reports during MI2 pretraining. For example, we find significant improvements for cardiomegaly and pleural thickening across all pretrain dataset sizes. We hypothesize that this effect is due to the implicit clinical context conveyed by the presence of medical tubes. Such devices often indicate severe illness and correlate with specific pathologies. Including tube presence labels may help the model distinguish between disease-related and device-induced visual features, with tube presence potentially acting as a proxy for disease severity. However, since we do not observe the same behavior in \Cref{fig:mayo_findings}, this may also reflect shortcut learning, likely due to the small dataset size \citep{perez-garcia_radedit_2025, geirhos_shortcut_2020}. Another issue related to limited data can be seen in \Cref{tab:vindr_findings_class}, where we compare models trained on the full INST-CXR-BENCH (3.5M samples) to open-weights models. Surprisingly, MI2, even when pretrained on 3.5M (image, report, tube presence label) samples, does not consistently outperform the RAD-DINO open-weights model, contrary to the trend in all of our other experiments, suggesting benchmark saturation.

\subsection{Lines and tubes segmentation on holdout dataset RANZCR-CLiP}
\label{sec:tubes_ranzcr}
\begin{figure}[h]
    \centering
    \begin{minipage}[h]{0.49\linewidth}
        \centering
        \includegraphics[width=\linewidth]{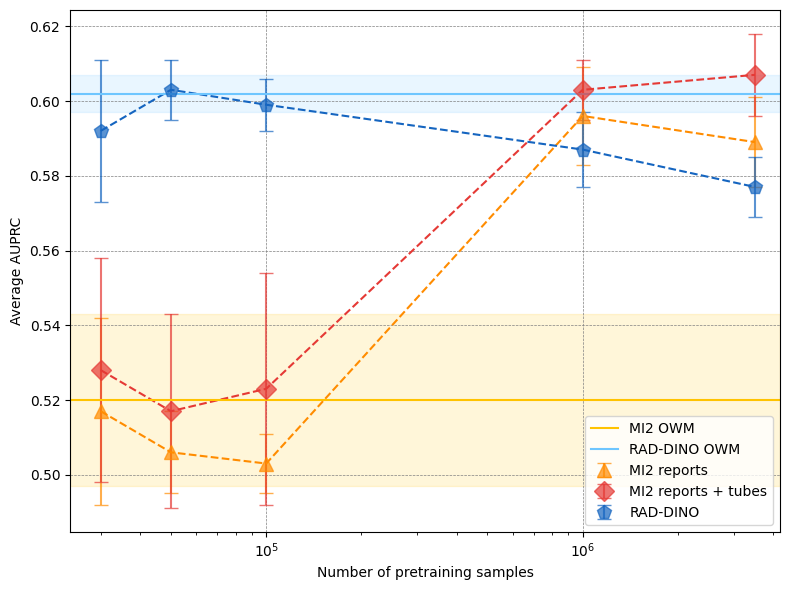}
    \end{minipage}
    \hfill
    \begin{minipage}[h]{0.49\linewidth}
        \centering
        \includegraphics[width=\linewidth]{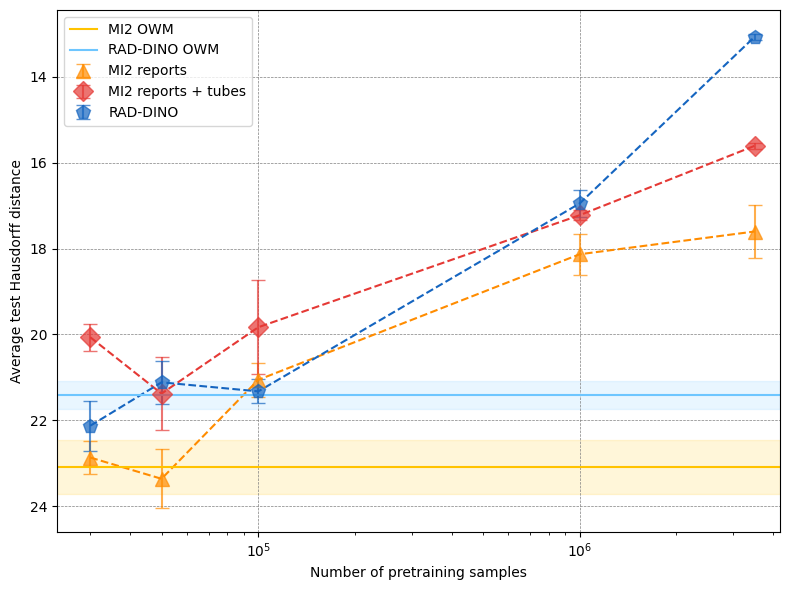}
    \end{minipage}
    \caption{Performance of encoders as a function of vision encoder pretraining with increasing sample sizes from INST-CXR-BENCH. Left: Findings classification on holdout dataset VinDR, AUPRC averaged across seven findings. Right: Lines and tubes segmentation on holdout dataset RANZCR-CLiP, Hausdorff distance averaged across four l\&t.}
    \label{fig:vindr_findings_and_ranzcr_hausdorff}
\end{figure}

Since INST-CXR-BENCH does not contain segmentation masks, we train and evaluate l\&t segmentation models on a public holdout dataset, called RANZCR-CLiP \citep{tang_clip_2021}, consisting of 17k CXR (75\%/25\% train/test). Note: While we are varying the number of pretraining samples, the amount of samples to train the segmentation model is always the same. We use the same feature pyramid architecture as in the classification setup, see \Cref{sec:classification}. For MI2, additional upsampling layers are applied to match the spatial resolution of RAD-DINO feature maps. A linear segmentation head followed by upsampling to the original image size (518$\times$518) is applied. All segmentation models are trained using Dice loss. For evaluating tube-like structures, we prioritize the Hausdorff distance as the primary metric due to its sensitivity to spatial localization. In \Cref{fig:ranzcr_dice}, we additionally provide the scaling curves for the DICE metric. In \Cref{fig:vindr_findings_and_ranzcr_hausdorff} (right), we see that RAD-DINO significantly outperforms MI2 when MI2 is pretrained using report-only supervision for all pretraining dataset sizes but 30k and 100k, which aligns with expectations for segmentation tasks \citep{bolya_perception_2025}. In general, scaling trends are inconsistent below 100k pretraining samples, likely due to the small size of the segmentation benchmark. However, incorporating tube presence labels during MI2 pretraining (via UniCL) leads to significant performance gains, surpassing the RAD-DINO open-weights model for all but one pretraining dataset size and closing the gap to the continually pretrained RAD-DINO across almost all pretraining dataset sizes. This suggests that the added labels provide valuable spatial context during contrastive learning. At the full 3.5M scale RAD-DINO significantly outperforms all other models. Extrapolating the performance trend of RAD-DINO pretrained on 100k, 1M, and 3.5M samples from INST-CXR-BENCH, we conclude that DINOv2 scales more effectively than CLIP/UniCL for l\&t segmentation tasks, which is in agreement with \citep{fan_scaling_2025}. Last, in \Cref{tab:ranzcr_segmentation}, we compare models trained on the full INST-CXR-BENCH dataset (3.5M samples) with open-weights models. RAD-DINO pretrained on 3.5M CXRs from INST-CXR-BENCH significantly outperforms all other models across all l\&t types.
\begin{table}[h]
\centering
\small
\caption{Findings classification on holdout dataset VinDR: Classification accuracy of open-weights encoder checkpoints and encoders pretrained with the full INST-CXR-BENCH (3.5M samples)}
\label{tab:vindr_findings_class}

\begin{tabular}{lllllllll}
\toprule
 Finding & AE & CM & LO & PE & PL-T & PF & TB & AVG \\
 Prevalence & 28.52\% & 23.63\% & 7.01\% & 8.28\% & 11.68\% & 13.71\% & 7.18\% \\
\midrule
MI2 OWM & 36.3\textsubscript{7.0} & 66.0\textsubscript{7.3} & 14.1\textsubscript{0.3} & \textbf{79.1\textsubscript{2.3}} & 40.2\textsubscript{2.6} & 57.4\textsubscript{4.5} & 71.3\textsubscript{5.5} & 52.0\textsubscript{2.3} \\
MI2 reports & \textbf{49.0\textsubscript{3.6}} & 78.6\textsubscript{1.4} & \textbf{16.0\textsubscript{1.7}} & 75.8\textsubscript{2.6} & 47.7\textsubscript{2.8} & \textbf{65.6\textsubscript{1.0}} & \textbf{79.8\textsubscript{0.8}} & \textbf{58.9\textsubscript{1.2}}\\
MI2 reports + tubes & \textbf{50.6\textsubscript{1.9}} & \textbf{81.4\textsubscript{1.0}} & \textbf{17.1\textsubscript{0.9}} & 77.0\textsubscript{4.7} & \textbf{52.6\textsubscript{0.7}} & \textbf{65.9\textsubscript{0.7}} & \textbf{80.4\textsubscript{0.5}} & \textbf{60.7\textsubscript{1.1}} \\
RAD-DINO OWM & \textbf{48.1\textsubscript{0.7}} & \textbf{83.2\textsubscript{0.4}} & \textbf{17.5\textsubscript{0.9}} & \textbf{83.5\textsubscript{0.1}} & 47.3\textsubscript{1.1} & \textbf{65.2\textsubscript{0.1}} & 76.6\textsubscript{0.9} & \textbf{60.2\textsubscript{0.5}} \\
RAD-DINO & 43.3\textsubscript{5.2} & 77.9\textsubscript{1.4} & \textbf{17.8\textsubscript{0.5}} & \textbf{80.5\textsubscript{0.6}} & 46.3\textsubscript{1.3} & \textbf{63.5\textsubscript{0.8}} & 74.7\textsubscript{0.7} & \textbf{57.7\textsubscript{0.8}} \\
\bottomrule
\end{tabular}

\centering
\small
AE: Aortic Enlargement, CM: Cardiomegaly, LO: Lung Opacity, PE: Pleural Effusion, PL-T: Pleural Thickening, PF: Pulmonary Fibrosis, TB: Tuberculosis, AVG: Mean of micro averaged AUPRC across 3 seeds.
\end{table}

\begin{table}[h]
\centering
\small
\caption{Lines and tubes segmentation on holdout dataset RANZCR-CLiP:  Hausdorff distance for segmentation with different open-weights models and encoders pretrained with the full INST-CXR-BENCH dataset (3.5M samples).}
\label{tab:ranzcr_segmentation}
\begin{tabular}{llllll}
\toprule
 Tube category & SGC & CVC & ETT & NGT & AVG \\
 Prevalence & 0.91\% & 51.54\% & 17.43\% & 18.49\% \\
\midrule
MI2 OWM & 39.6\textsubscript{0.7} & 73.5\textsubscript{1.4} & 23.1\textsubscript{0.6} & 80.0\textsubscript{1.7} & 23.1\textsubscript{0.6} \\
MI2 reports & 39.4\textsubscript{0.2} & 48.3\textsubscript{0.6} & 17.6\textsubscript{0.6} & 58.9\textsubscript{0.8} & 17.6\textsubscript{0.6} \\
MI2 reports + tubes & 40.5\textsubscript{0.3} & 40.0\textsubscript{0.3} & 15.6\textsubscript{0.1} & 49.8\textsubscript{0.9} & 15.6\textsubscript{0.1} \\
RAD-DINO OWM & 39.0\textsubscript{0.2} & 35.3\textsubscript{0.2} & 21.4\textsubscript{0.3} & 38.9\textsubscript{0.6} & 21.4\textsubscript{0.3} \\
                    RAD-DINO & \textbf{38.2}\textsubscript{0.1} & \textbf{26.9}\textsubscript{0.3} & \textbf{13.1}\textsubscript{0.1} & \textbf{24.8}\textsubscript{0.2} & \textbf{13.1}\textsubscript{0.1} \\
\bottomrule
\end{tabular}

\centering
\small
SGC: Swan Ganz Catheter, CVC: Central Venous Catheter, ETT: Endotracheal Tube, NGT: Nasogastric Tube, AVG: Mean of micro averaged Hausdorff Distance across 3 seeds.
\end{table}
\subsection{Report generation on INST-CXR-BENCH-REPORT-GEN}
\label{sec:maira_exps}
\begin{figure}[h]
    \centering
    \begin{minipage}[h]{0.49\linewidth}
        \centering
        \includegraphics[width=\linewidth]{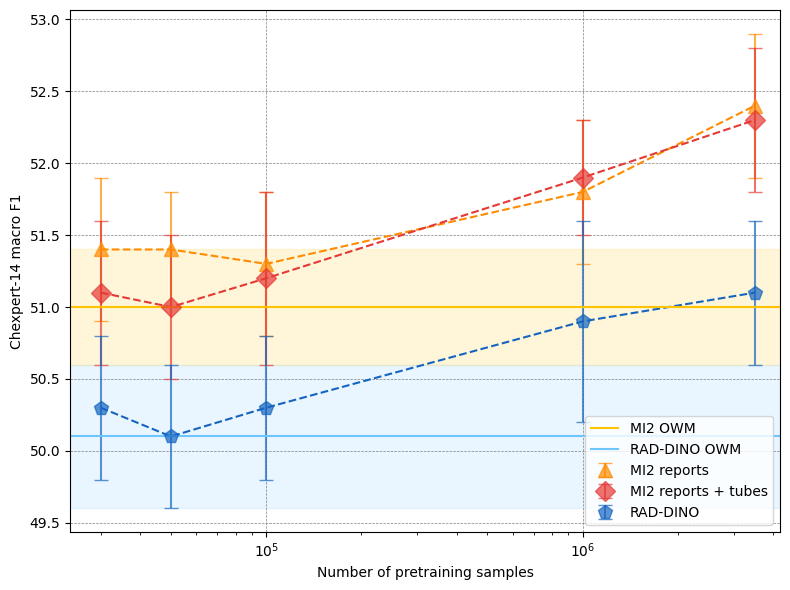}
    \end{minipage}
    \hfill
    \begin{minipage}[h]{0.49\linewidth}
        \centering
        \includegraphics[width=\linewidth]{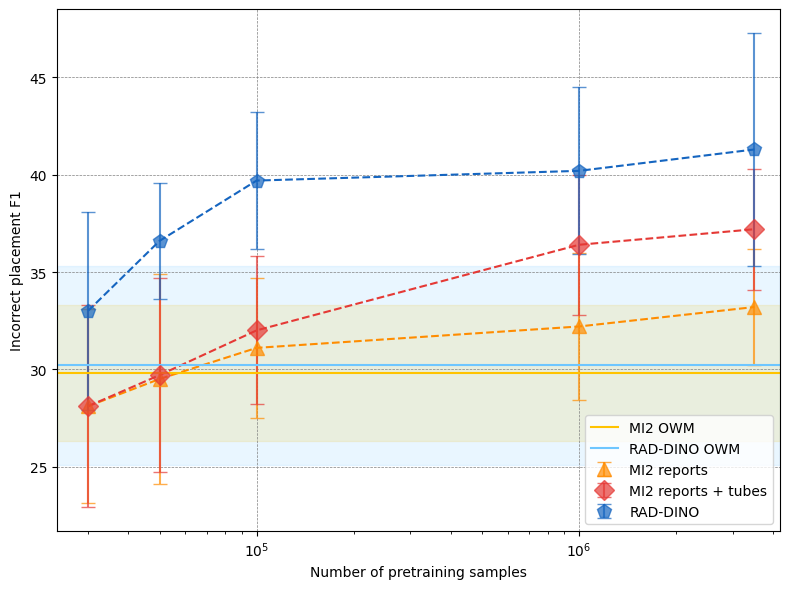}
    \end{minipage}
    \caption{Report generation performance on INST-CXR-BENCH-REPORT-GEN as a function of vision encoder pretraining with increasing sample sizes INST-CXR-BENCH. Left: CheXbert Macro F1 averaged across 14 findings. Right: Incorrect Placement F1.}
    \label{fig:mayo_findings_generation}
\end{figure}

We compare three vision encoder pretraining approaches: RAD-DINO, MI2 reports, and MI2 reports + tube labels for report generation using the MAIRA-2 13B framework \citep{hyland_maira-1_2024, bannur_maira-2_2024}. Training and evaluation use 2.5M studies from INST-CXR-BENCH (subset: INST-CXR-BENCH-REPORT-GEN). Inputs include current frontal and lateral views, prior frontal view, clinical indication, comparison sections, and the full prior report. While encoder pretraining size varies, the report generation training set remains fixed. Unlike earlier experiments (\Cref{sec:classification}, \ref{sec:tubes_ranzcr}), we train one MAIRA-2 model per pretraining size and method due to the high cost of training a 13B LLM. To still measure experiment variability, the results in \Cref{fig:mayo_findings_generation} and \Cref{tab:mayo_findings_generation} report medians and 95\% CIs from 500 bootstrap samples. The performance of MAIRA-2 is assessed using four metrics for natural language generation (NLG) and clinical efficacy (CE): ROUGE-L (NLG) \citep{lin_automatic_2004}, CheXbert Macro F1-14 (CE) \citep{smit_chexbert_2020}, RadFact Logical F1 (CE) \citep{bannur_maira-2_2024}, and a novel CE metric, called Incorrect Placement F1, measuring detection of misplaced lines/tubes, a clinically critical task requiring curve-continuity features (\Cref{sec:raddino}). Due to compute-heavy metrics (especially CheXbert and RadFact), inference uses a 40k-study subset, each study retaining one frontal, one lateral, and one prior frontal image per case. The resulting test dataset is designed to reflect the real-world distribution of l\&t encountered in an ICU setting. Given the high cost of MAIRA-based generation, we ran ablations to find the best layer(s) to extract features. We compared the four-layer combination from \Cref{sec:classification} with final-layer features and found the latter performed best for both RAD-DINO and MI2. Notably, MI2’s last layer uses only 289 image tokens versus 1369 for RAD-DINO, reducing training time by 75\% and inference time by 33\%. Across our experiments, the largest performance differences among pretraining methods appear in CheXbert Macro F1-14 and Incorrect Placement F1 (\Cref{fig:mayo_findings_generation}). Consistent with Section \ref{sec:classification}, MI2 significantly outperforms RAD-DINO on CheXbert Macro F1-14, which strongly correlates with findings classification. Notably, performance curves for all three strategies show parallel trends, suggesting a shared scaling exponent $k$ but different scaling multipliers $\alpha$. As expected, the two MI2 variants (with and without tube-label supervision) show no significant difference across scales. Compared to open-weight models, both RAD-DINO and MI2 only surpass baseline performance when trained on 1M or more examples. For Incorrect Placement F1, RAD-DINO significantly outperforms MI2 trained solely with report supervision, consistent with \Cref{sec:tubes_ranzcr}. We attribute this to radiology reports often lacking detailed descriptions of l\&t placements, including tip positions. However, adding explicit tube presence labels to MI2 pretraining narrows the gap. Unlike the near-linear scaling seen with CheXbert Macro F1-14, Incorrect Placement F1 saturates quickly, likely due to the low ($\sim$12\%) prevalence of incorrectly placed tubes in INST-CXR-BENCH-REPORT-GEN. For this task, RAD-DINO exceeds open-weights models with just 100k samples, whereas MI2 requires $\geq$1M to consistently outperform open weights. In \Cref{tab:mayo_findings_generation}, we compare models trained on all INST-CXR-BENCH data (3.5M samples) to open weights. For ROUGE-L, CheXbert Macro F1-14, and Incorrect Placement F1, models pretrained on INST-CXR-BENCH significantly outperform their open-weight counterparts. For RadFact, both RAD-DINO versions are on par. For ROUGE-L, CheXbert Macro F1-14, and RadFact, MI2 outperforms RAD-DINO, likely because these metrics correlate with findings classification. Only for Incorrect Placement F1 do RAD-DINO models, both open-weight and continually pretrained, outperform MI2.
\begin{table}[h]
\centering
\small
\caption{Findings generation on INST-CXR-BENCH-REPORT-GEN dataset. Comparison of the open-weights models and encoders pretrained with the full INST-CXR-BENCH dataset (3.5M samples).}
\label{tab:mayo_findings_generation}

\begin{tabular}{lcccc}
\toprule
 & ROUGE-L & CheXbert Macro F1-14 & Incorrect Placement F1 & RadFact: logical F1 \\
\midrule
RAD-DINO OWM & 38.5 [38.2, 38.7] & 50.1 [49.6, 50.6] & 30.2 [26.0, 35.3] & 63.3 [63.1, 63.6] \\
RAD-DINO & 39.0 [38.8, 39.2] & 51.1 [50.6, 51.6] & \textbf{41.3 [36.1, 47.3]} & 63.0 [62.8, 63.2] \\
MI2 OWM & 38.7 [38.5, 38.9] & 51.0 [50.5, 51.4] & 29.8 [26.0, 33.3] & 63.8 [63.6, 64.0] \\
MI2 reports & \textbf{39.5 [39.3, 39.7]} & \textbf{52.4 [51.9, 52.9]} & 33.2 [30.2, 36.3] & \textbf{64.7 [64.5, 64.9]} \\
MI2 reports + tubes & \textbf{39.5 [39.3, 39.8]} & \textbf{52.3 [51.8, 52.8]} & 37.2 [33.7, 40.3] & \textbf{64.8 [64.6, 65.0]} \\
\bottomrule
\end{tabular}
\end{table}
\section{Conclusion}
Our study demonstrates that continual pretraining of open-weight models on large-scale CXR datasets yields significantly improved vision encoders. This highlights the promise of developing center-specific foundation models, allowing large medical institutions to tailor encoders to their unique patient populations and imaging protocols. We establish clear scaling laws up to 3.5M samples, indicating that existing foundation models such as MedImageInsight and RAD-DINO continue to benefit from additional data. Notably, despite using the same pretraining data and compute budget, these models exhibit complementary strengths: MedImageInsight (CLIP-style) excels at findings-related tasks, whereas RAD-DINO (DINOv2-style) performs better on tube-related tasks. Moreover, incorporating tube presence labels into MedImageInsight pretraining via UniCL closes the performance gap with RAD-DINO, underscoring the importance of structured supervision, even at scale. This demonstrates the value of structured labels extracted by LLMs such as GPT; see \Cref{sec:gpt_extraction} for a detailed discussion. In our experiments, however, many scaling curves deviate from idealized power-law behavior. In the small-data regime, performance is often noisy, while in the large-data regime, improvements can plateau. Domain shift (e.g., training on data from different hospitals) further complicates these trends. The results in \Cref{sec:findings_vindr} especially highlight the need for larger, multi-center benchmark datasets to effectively compare CXR vision encoders. Overall, our findings suggest that continual pretraining of MI2 using the UniCL framework, combined with automated label extraction, is the most effective strategy for medical centers aiming to train foundation vision encoders on their own data. We also emphasize the importance of a large and diverse test dataset, diverse both in tasks and metadata, to thoroughly evaluate the performance of pretrained vision encoders. Finally, the scaling curves in \Cref{fig:mayo_findings} and \Cref{fig:mayo_tubes} indicate that improving the average performance of an vision encoder may require billions of training samples. However, average performance can be misleading, as it aggregates tasks that are nearly saturated with those that could benefit significantly from additional data. This underscores that a brute-force approach of simply collecting more data is not an effective path forward. Instead, efforts should focus on identifying and prioritizing underrepresented or low-performing tasks, potentially through data selection strategies, like active learning \citep{ren_survey_2021} and data filtering \citep{vo_automatic_2024, mindermann_prioritized_2022}.


\subsubsection*{Acknowledgments}
This work is supported in part by the generosity of Stephen A. and Linda L. Odell.

\bibliography{iclr2026_conference}
\bibliographystyle{iclr2026_conference}

\newpage
\appendix
\section{Appendix}
\subsection{Comparison of MI2 and RAD-DINO}
\begin{table}[H]
\small
\caption{Comparison of MedImageInsight (MI2) and RAD-DINO}
\label{tab:comparison}
\centering
\begin{tabular}{>{\raggedright\arraybackslash}p{4cm}>{\raggedright\arraybackslash}p{5.5cm}>{\raggedright\arraybackslash}p{5.5cm}}
\hline
\textbf{} & \textbf{MedImageInsight (MI2)} & \textbf{RAD-DINO} \\
\hline
\textbf{Architecture / \#parameters} & DAViT / 360M & ViT-B / 87M \\
\hline
\textbf{Training method} & UniCL & Dinov2 \\
\hline
\textbf{Training data} & CXRs (500k) + other modalities (3.3M) & CXRs (800k) \\
\hline
\multirow{4}{=}{\textbf{\# of tokens (518×518 input image size)}}& Block0: 130×130 = 16,900 & \multirow{4}{=}{Everywhere: 37×37 = 1,369} \\
& Block1: 65×65 = 4,225 & \\
& Block2: 33×33 = 1,098 & \\
& Block3: 17×17 = 289 & \\
\hline
\multirow{4}{=}{\textbf{Token dimension}} 
& Block0: 256 & \multirow{4}{=}{Everywhere: 768} \\
& Block1: 512 & \\
& Block2: 1024 & \\
& Block3: 2048 & \\
\hline
\end{tabular}
\end{table}

\subsection{Lines and tubes segmentation on holdout dataset RANZCR-CLiP}
In \Cref{tab:ranzcr_segmentation}, we compare models trained on the full INST-CXR-BENCH dataset (3.5M samples) with open-weights models.
\begin{figure}[H]
    \centering
    \includegraphics[width=0.5\linewidth]{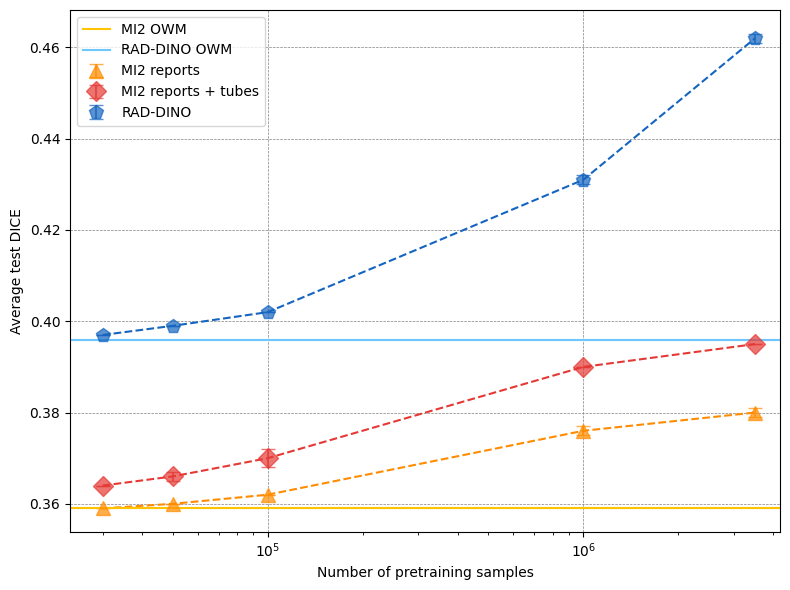}
    \caption{Lines and tubes segmentation performance on RANZCR-CLiP as a function of vision encoder pretraining with increasing sample sizes from INST-CXR-BENCH. DICE averaged across four l\&t.}
    \label{fig:ranzcr_dice}
\end{figure}

\subsection{Label extraction}
\label{sec:gpt_extraction}
We implemented an LLM-based pipeline to extract findings labels as well as l\&t labels from CXR radiology reports. This involved the use of detailed prompts that were engineered with the help of radiologists and iteratively refined for accurate label extraction. We used GPT-4o \citep{openai_gpt-4o_2024} endpoints for extracting the findings and l\&t labels. This process was designed to support structured data generation for downstream clinical applications, with an emphasis on consistency, reliability, and alignment with radiologist expectations. We extracted the presence or absence of 19 findings: hyperinflation, interstitial lung disease pattern, atelectasis, costophrenic angle blunting, pleural effusion, pneumothorax, adenopathy, enlarged pulmonary artery, arterial calcification, osseous abnormalities, rib fracture, bronchial wall thickening, hernia, subcutaneous air/emphysema, opacity, vascular congestion, cardiomegaly, diaphragm elevation, pulmonary edema. The findings were chosen to cover a wide range of appearances (some are more diffuse/texture like, others are more localized/shape like) and areas of a CXR (from the esophagus to the diaphragm), and have a good support (at least 10k in the test set) in the dataset. There are ten l\&t types whose presence or absence we extracted: Internal Jugular Central Venous Catheter (CVC), Peripherally Inserted Central Catheter, Subclavian CVC or Port-A-Cath, Endotracheal Tube, Tracheostomy Tube, Nasogastric Tube, Swan-Ganz Catheter, Chest Tube, Mediastinal Drain, and Intra-Aortic Balloon Pump. These are the most common l\&t devices observed in CXR reporting in practice, and radiologists considered them important for analysis. We evaluated the accuracy of the GPT-based l\&t labels extraction using a manually annotated hold-out dataset of 115 samples and achieved an F1-score of 0.94. This manual evaluation provided key insights into error modes and prompted refinements to both the input formatting and the GPT prompt design. What follows is an example prompt for extracting chest tube labels:

\textcolor{gray}{\texttt{You are an AI radiology assistant. You are helping to process reports for Chest X-rays by extracting information about lines and tubes visible in the image, by looking at the reports. In radiology reports, "left" corresponds to the left side of the patient, which is the right side of the X-ray; similarly "right" corresponds to the right side of the patient, which is the left side of the X-ray; use the same terminology.}}

\textcolor{gray}{\texttt{You will be given the report for the current study (marked by "Current Study") which describes the findings from the chest X-ray(s) taken at the that time. Each report will have the date of the report, the reason for exam, and the impression, which contains the radiologist's observations.}}

\textcolor{gray}{\texttt{The goal is to use the reports to extract information about lines and tubes which can be seen in the current X-ray.
Look at current report for the specified line/tube and its side. Check if the specified line/tube is mentioned.
Check if the current report states if the line/tube is correctly placed or indicates any malpositioning (for instance, doubled up, looped, kinked, coiled), and should be repositioned or retracted.}}
\textcolor{gray}{\texttt{Only extract lines and tubes mentioned in the current report. Only describe changes which are described in the current report.}}

\textcolor{gray}{\texttt{Extract information in JSON format as a list of each line/tube visible in the current X-ray image. Each line/tube should have a single entry. There can be multiple types of lines/tubes in the report, as well as multiple instances of the same type or even the same subtype; in all cases, ensure that each one has a separate entry in the JSON list. If there are no lines/tubes then output an empty list.}}

\textcolor{gray}{\texttt{\# JSON entry fields}}

\textcolor{gray}{\texttt{- reference\_sentence (this should contain the original sentence, sub-sentence, or multiple sentences from the report describing all details about the line/tube)
- type: the line/tube type exactly as written in the report
- type\_categorical: the line/tube type formatted to fall into one of a fixed number of categories that will be defined later.
- placement: if described in the report, whether the line/tube is correctly placed or incorrectly placed (correct or incorrect). If it is not explicitly described, use the tip location to infer the placement, that will be defined later. If it is described but it’s unclear what category it falls into, write “unclear”. Otherwise N/A.}}

\textcolor{gray}{\texttt{\# Lines and tubes to extract}}

\textcolor{gray}{\texttt{In this pass, only extract information about chest tubes. Chest tubes are inserted through the chest wall into the pleural space and are used to drain fluid, blood, or air.
There are other ways to describe a chest tube including chest drain, pleural drain, pleural catheter, pigtail pleural drain, pigtail catheter, drainage catheter, drainage tube, thoracostomy tube, PleurX catheter, etc.
Different terms may be used in different reports; use in such cases, if there are multiple chest tubes and it is ambiguous which one corresponds to which in previous reports, use information about insertion side and tip location to determine which are which. If chest tubes are described as bilateral or bibasilar etc., means that more than one chest tubes are present in both sides of the chest i.e. there is one on each side of the body, then output two entries, one for side\_categorical left, and the other for side\_categorical right.}}

\textcolor{gray}{\texttt{\#\# Additional information
Do not confuse chest tubes with mediastinal drains and pericardial drains, which are inserted in the mediastinum rather than the pleural space. Also do not confuse chest tubes with any other kinds of tubes such as feeding tubes, tracheostomy tubes, or endotracheal tubes.}}

\textcolor{gray}{\texttt{It is common for there to be multiple chest tubes in place at one time. Remember that each each individual chest tube must have a separate entry in the output list.}}

\textcolor{gray}{\texttt{\#\# Placement Information
For the placement field:
Write "incorrect" if that line/tube is described as misplaced or malpositioned (e.g. kinked, coiled, doubled up) and/or should be repositioned or withdrawn.
Write "incorrect" if the report mentions a pleural effusion or pneumothorax on the same side as the chest tube that is at least moderate in size or larger/worsened than before.
Write "incorrect" if the tube or side port is outside of the chest cavity.
Write "correct" if the current report describes a "stable position" of that line/tube or that line/tube being "in place".
If correct/incorrect placement is not explicitly described in the report, use the following mapping from the extracted tip location:
{'upper': 'correct', 'lower': 'correct', 'middle': 'correct', 'below diaphragm': 'incorrect', 'side port outside rib cage': 'incorrect', 'outside chest': 'incorrect', 'adjacent to mediastinum/esp aorta': 'incorrect', 'unclear': 'unclear', 'N/A': 'N/A'}
If tip location is described but placement can't be inferred from the above mapping, write "unclear".
Write "N/A" if there is no tip location or placement information about that line/tube in the report.
Write "N/A" if the current report describes that line/tube as having been removed.}}

What follows is an example prompt for extracting findings from CXR reports:

\textcolor{gray}{\texttt{You are an AI radiology assistant. You are helping process reports from chest X-rays.
In radiology reports, “left” corresponds to the left side of the patient, which is the right side of the X-ray; similarly, “right” corresponds to the right side of the patient, which is the left side of the X-ray.
Each radiology report contains several sections, such as the findings, impression, comparison, indication, and technique sections.}}

\textcolor{gray}{\texttt{Please extract information about all the findings and diseases from the radiology report that refer to findings visible in a chest X-ray or disease diagnosed from a chest X-ray, and categorize certain elements. 
Your task is to extract information about all findings and diseases from the current report and prior structured reports (if available) in JSON format as a list of dictionaries.
**If a finding or disease is present in the prior structured report but not in the current report, ensure it is included in the current report output with all of its details from the prior report.** 
Each unique combination of finding/disease and region should have a single entry, carrying forward all prior information as needed.}}

\textcolor{gray}{\texttt{Each entry should use the keys given below:
"finding\_type": The finding type, value should be either DISEASE or FINDING. 
FINDING represents an observation in the chest x-ray. DISEASE represents the interpretation or diagnosis from the observations in the chest x-ray.
Return the finding\_type value depending on which list the extracted "label" below belongs to. Return DISEASE if label is in DISEASE list or FINDING if label is in FINDING list.}}

\textcolor{gray}{\texttt{We will use the word "finding" in the rest of the prompt, to represent a FINDING or a DISEASE.}}

\textcolor{gray}{\texttt{"reference\_phrase": The phrase associated with the finding. Make sure to provide the exact phrase from the report. Don't change the phrase at all, extract it as it is. If the same finding is present in the current report, update it with the new phrase. Otherwise, retain the phrase from the prior report.}}

\textcolor{gray}{\texttt{"label" : The finding label mentioned in the phrase. The value must come from the provided list of DISEASE or FINDING.
Provide the value "No finding" when a phrase mentions anatomical structures with normal observations. For example: "The cardiomediastinal silhouette is normal", "The imaged upper abdomen is unremarkable", "Lungs are clear", "Pulmonary vasculature is normal", "The cardiomediastinal silhouette is within normal limits", "The cardiac, mediastinal and hilar contours are normal".}}

\textcolor{gray}{\texttt{DISEASE:
\$DISEASE}}

\textcolor{gray}{\texttt{FINDING:
\$FINDINGS}}

\textcolor{gray}{\texttt{**Please note:**
1. 'Pulmonary vascular engorgement' and 'Vascular engorgement' are other ways of refering to Pulmonary venous hypertension.
2. 'Mediastinal widening' and 'Enlarged cardiomedistinum' are other ways of refering to Enlarged cardiomediastinum.
3. 'Hyperaeration' and 'Overinflation' are other ways of refering to Hyperinflation.
4. 'Negative chest, 'chest negative' and 'no acute disease in the chest' are other ways of refering to No finding.
5. 'Prosthetic valve' is another way of refering to Valve prosthesis.
6. 'Enlarged cardiomegaly' is another way of refering to Cardiomegaly.
7. 'Fibrosis' is another way of refering to Pulmonary fibrosis.
8. 'Pulmonary opacity' is another way of refering to lung opacity.
9. Only when 'linear' when used with 'fibrosis' i.e. 'linear fibrosis' is another way of refering to scarring.
10.'Infiltration' is another way of refering to infiltrate.}}

\textcolor{gray}{\texttt{**Instruction for handling out of list values**
**You should strictly stick to FINDING and DISEASE labels for "label" category.**
**When you find a value of a category which is not from one of the given values for that category (except for "label" category), assign "Other" to it. If the category doesn't exist in the phrase, assign "N/A" to it.**}}

\textcolor{gray}{\texttt{**Format of each output structured finding**:
$[\{$
    "finding\_type" : "",
    "reference\_phrase": "",
    "label": "",
$\}$
$]$}}

\textcolor{gray}{\texttt{**Instructions for Handling All Findings**
1. If the same finding is mentioned in multiple sections with different regions, comparison status, is\_positive status, severity, anatomy, morphology, spatial distribution or spatial comparison, then extract each instance separately.
2. If the finding is present in multiple phrases, return multiple JSON items for each finding separately.
3. Include normal or negative findings as well. If a finding is negative give the label for that and mark is\_positive as "No" for it. e.g: labels present in the phrase : "There is no atelectasis or lung opacity seen."  are $[$'Atelectasis', 'Lung Opacity'$]$
4. Match the finding sentences first from the current report with the prior structured report phrases, then create the final structured report.}} 

\textcolor{gray}{\texttt{**Only if the input has prior report incorporate the below changes otherwise ignore.**}}

\textcolor{gray}{\texttt{**Instructions for Incorporating Prior Findings**:
1. **Inclusion of Previous Findings**: All findings, including negative findings, from the prior structured report should be included in the current report output, even if they are not mentioned in the current report. If no prior structured report is provided, treat the current report as the first report for the patient.
2. **Current Report First**: In the final structured report, if there is a prior structured report, give the current report phrases first and then the prior report phrases.
3. **Finding Propagation**: Propagate all findings with their corresponding values from the prior report unless new values are provided in the current report.
4. **Unchanged Findings**: If the current report does not specify changes in a finding, it should retain its values from the prior report in the output.
5. **Updates to Prior Findings**: If the current report updates an existing finding replace the previous values with the updated values for that finding.}}

\textcolor{gray}{\texttt{Don't provide any explanations.}}

\subsection{Significance test}
\label{sec:signify_test}
We compare two binary classifiers across multiple tasks and multiple random seeds using a hierarchical paired bootstrap procedure to estimate the difference in performance and its uncertainty. For each task, we collect the ground-truth labels and the predicted probabilities from both models across all seeds.
We begin performing stratified bootstrap resampling (500 bootstrap samples) of the test set for each task to preserve the original class balance. For every bootstrap replicate, we compute the AUPRC for each seed of both models using the resampled data. These seed-level metrics are then avaraged within each model. The difference between the aggregated metrics of the two models is saved for that replicate.
Repeating this process across many bootstrap replicates produces a distribution of differences for each task. From this distribution, we report the average difference and construct a percentile-based confidence interval at the 95\% level. To obtain an overall comparison across all tasks, we pool the bootstrap differences from every task into a single distribution (micro AUPRC) and compute the overall mean difference and its confidence interval.

\subsection{Metadata stratification}
\label{sec:metadata}

To evaluate potential biases and subgroup performance disparities, we stratify model performance across five metadata variables: ethnicity, sex, age, scanner manufacturer, and patient type (inpatient vs. outpatient). For each variable, we compare the best-performing MI2 and RAD-DINO models (trained with 3.5M INST-CXR-BENCH samples) with their corresponding paper checkpoints. We observe that performance trends are largely consistent across MI2 and RAD-DINO; that is, subgroups where MI2 underperforms tend to also show lower performance for RAD-DINO. For all models we are focusing on the findings classification task from \Cref{sec:finginds_internal}.

For the ethnicity metadata variable, we stratify the test set into two categories: White (87\%) and Non-White (13\%). A performance drop of 3\% is observed for the Non-White group. This is expected given the reduced sample size in this group. In addition, we assess performance by patient care setting. We find that models perform 5\% worse on outpatient scans, with a notably higher standard deviation across runs. We hypothesize that the drop in performance stems from a greater variability in outpatient imaging protocols and patient conditions.
\begin{figure}[H]
    \centering
    \begin{minipage}[h]{0.49\linewidth}
        \centering
        \includegraphics[width=\linewidth]{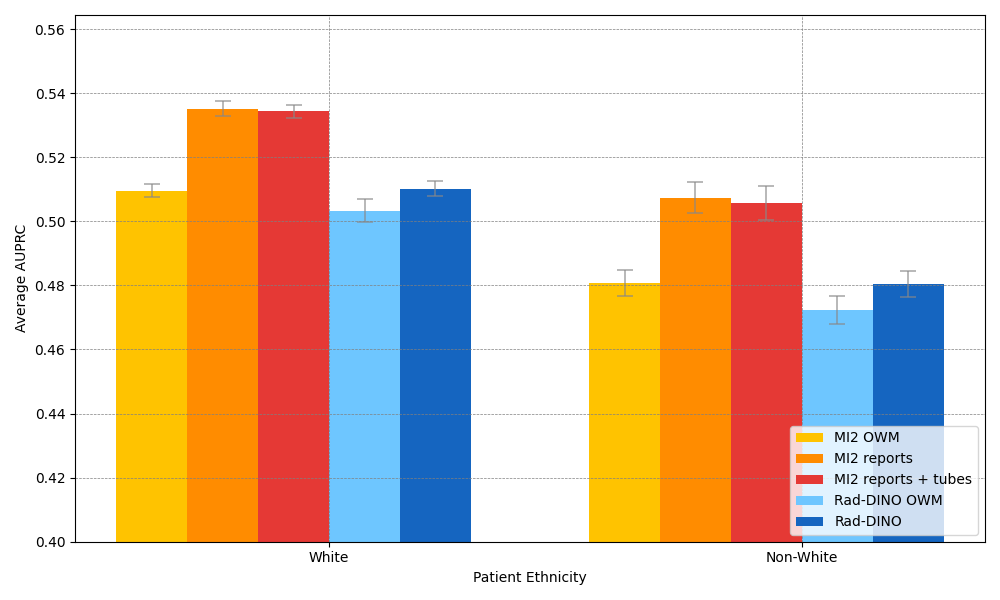}
    \end{minipage}
    \hfill
    \begin{minipage}[h]{0.49\linewidth}
        \centering
        \includegraphics[width=\linewidth]{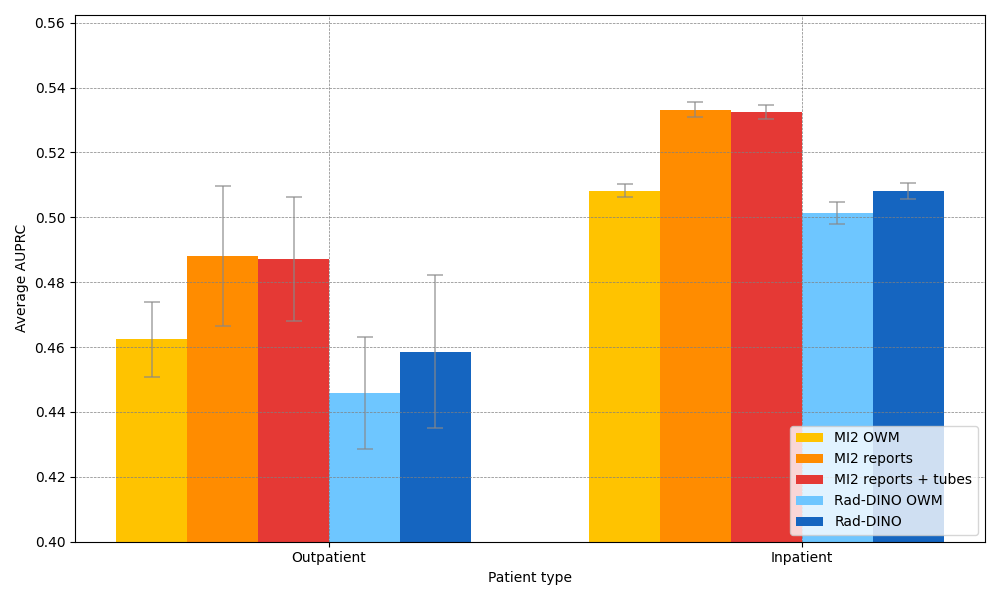}
    \end{minipage}
    \caption{Metadata stratification findings classification on INST-CXR-BENCH. Left: Average AUPRC across 20 findings stratified by ethnicity. Right: Average AUPRC across 20 findings stratified by patient type.}
    \label{fig:metadata_ethnicity_patient_type}
\end{figure}

We divide age into six groups: '20-30' (7\%), '30-40' (9\%), '40-50' (13\%), ‘50-60’ (22\%), '60-70' (22\%), '70-80' (17\%). A decrease in performance (5\%) is observed in the two youngest age groups, which also represents the smallest proportion of the dataset. 
When stratifying by sex (Female (48\%) and Male (49\%)) we find that both MI2 and RAD-DINO models perform slightly better for female patients, with an average performance increase of 2\%. We focus on the six most prevalent scanner manufacturers in the dataset. FUJIFILM Corporation (38\%), Carestream Health (24\%), GE Healthcare (15\%), SIEMENS (8\%), Philips (6\%). Among these, we observe a 6\% performance drop for scans from FUJIFILM Corporation compared to the best-performing group, Carestream Health. Performance on FUJIFILM scanners is likely worse because some original images are mislabeled as ‘derived’ in the DICOM tags. Since our subsetting prioritizes the latest original image (or derived if no original exists), this mislabeling may have caused FUJIFILM cases to rely on suboptimal images.
\begin{figure}[H]
    \centering
    \begin{minipage}[h]{0.49\linewidth}
        \centering
        \includegraphics[width=\linewidth]{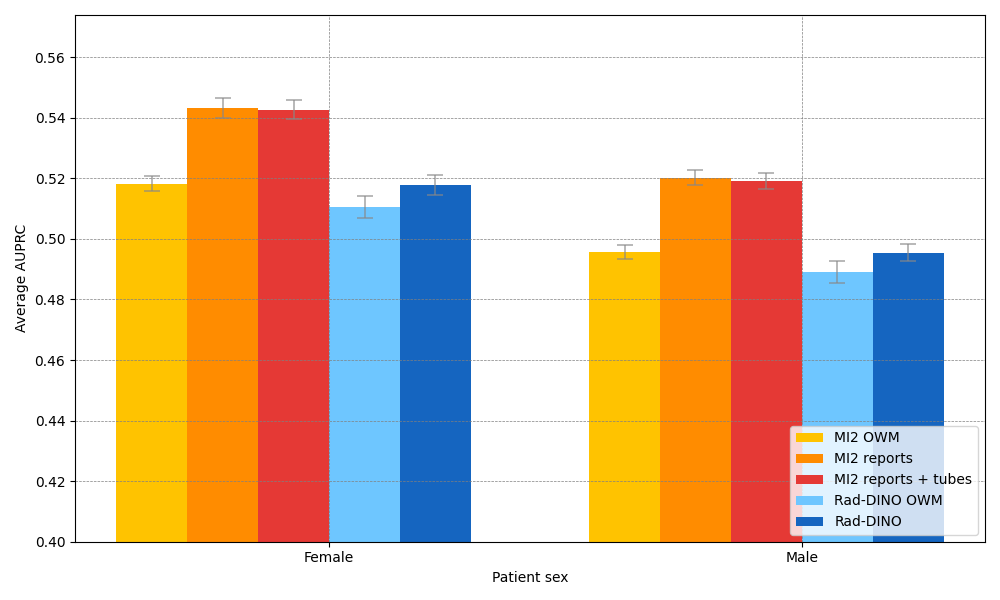}
    \end{minipage}
    \hfill
    \begin{minipage}[h]{0.49\linewidth}
        \centering
        \includegraphics[width=\linewidth]{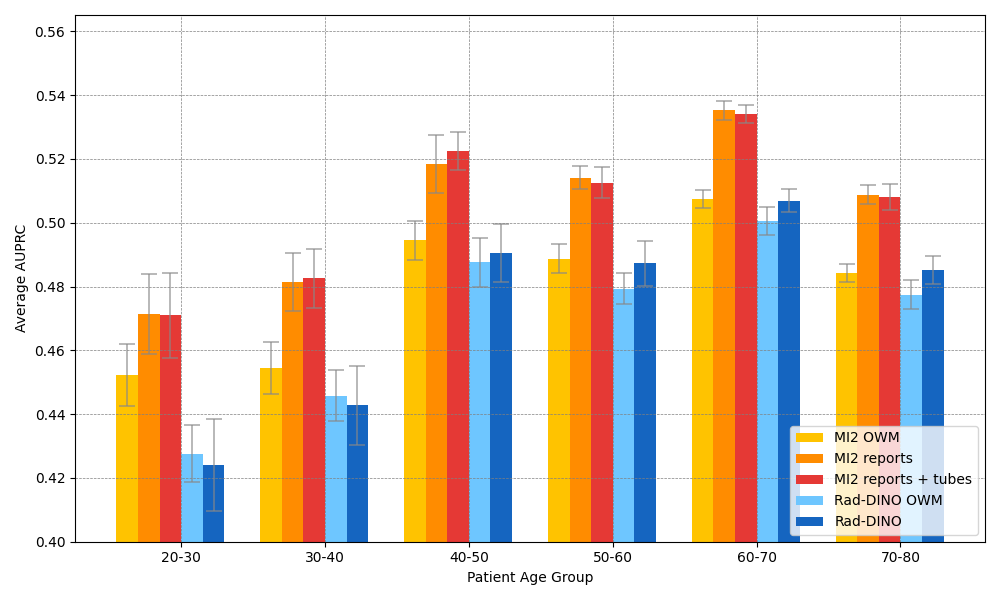}
    \end{minipage}

    \begin{minipage}[h]{0.6\linewidth}
        \centering
        \includegraphics[width=\linewidth]{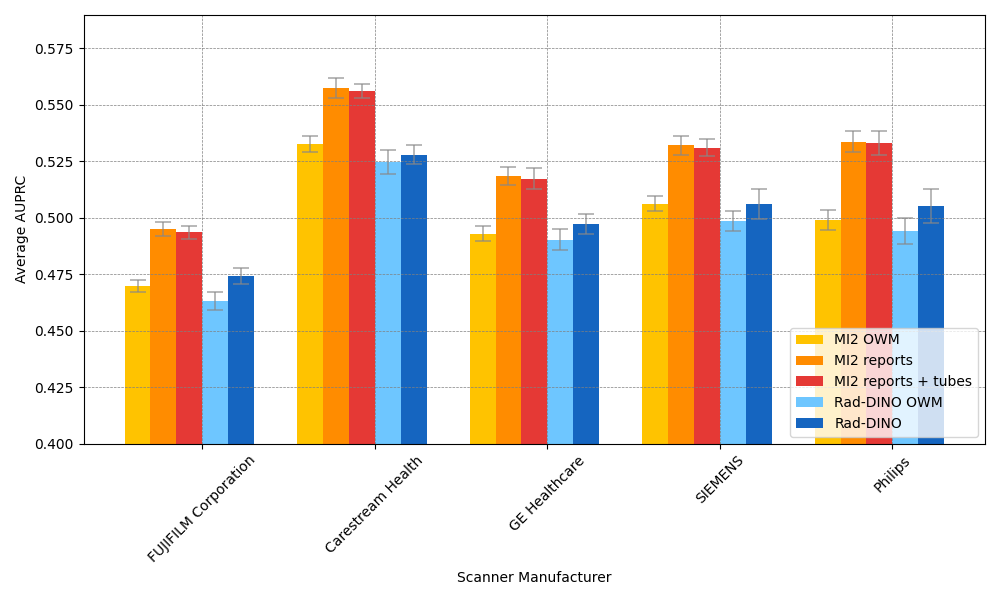}
    \end{minipage}
    \caption{Metadata stratification findings classification on INST-CXR-BENCH. Top left: Average AUPRC across 20 findings stratified by sex. Top right: Average AUPRC across 20 findings stratified by age. Bottom:  Average AUPRC across 20 findings stratified by manufacturer.}
    \label{fig:third_plot}
    \label{fig:metadata_scanner_age_sex}
\end{figure}

\end{document}

%% file: math_commands.tex
\usepackage{amsmath,amsfonts,bm}

\def\eqref#1{equation~\ref{#1}}

\def\1{\bm{1}}

\DeclareMathAlphabet{\mathsfit}{\encodingdefault}{\sfdefault}{m}{sl}
\SetMathAlphabet{\mathsfit}{bold}{\encodingdefault}{\sfdefault}{bx}{n}

